\pdfoutput=1

\documentclass[final,11pt]{article}

\usepackage{acl}
\usepackage{multirow}
\usepackage{times}
\usepackage{latexsym}
\usepackage{graphicx}
\usepackage{amssymb}
\usepackage{bm}
\usepackage{tikz}
\usepackage[utf8]{inputenc}
\usepackage{CJKutf8}
\usepackage{amsmath}
\usepackage{multirow}
\usepackage{booktabs}
\usepackage{enumitem}
\usepackage{longtable}
\usepackage{colortbl}
\usepackage{pifont}
\usepackage{fancyvrb}
\usepackage{xcolor}
\usepackage{algorithm}
\usepackage{algpseudocode}
\usepackage{multirow} 
\usepackage{booktabs}
\usepackage{enumitem}
\usepackage[T1]{fontenc}
\usepackage{hyperref} 

\usepackage[utf8]{inputenc}

\usepackage{microtype}

\usepackage{inconsolata}

\usepackage{graphicx}
\usepackage{listings}
\title{Enhancing LLM Language Adaption through Cross-lingual In-Context Pre-training}

\author{Linjuan Wu$^{1}$\footnotemark[1], Haoran Wei$^{2}$\footnotemark[2], Huan Lin$^{2}$, Tianhao Li$^{2}$, Baosong Yang$^{2}$, Fei Huang$^{2}$\footnotemark[4], {\bf Weiming Lu$^{1,}$\footnotemark[3]}\\
        $^{1}$Zhejiang University\\
        $^{2}$Tongyi Lab, Alibaba Group\\
        $^{1}$\texttt{\{wulinjuan525,luwm\}@zju.edu.cn}\\
        $^{2}$\texttt{\{funan.whr, lilai.lh, chongsheng.lth, yangbaosong.ybs, f.huang\}@alibaba-inc.com}}

\begin{document}
\maketitle

\renewcommand{\thefootnote}{\fnsymbol{footnote}}
\footnotetext[1]{Work done during internship at Tongyi Lab.}
\footnotetext[2]{Contributed equally.}
\footnotetext[3]{Corresponding authors.}
\footnotetext[4]{Google Scholar ID is \href{https://scholar.google.com/citations?user=9r98PpoAAAAJ}{9r98PpoAAAAJ}}

\begin{abstract}

Large language models (LLMs) exhibit remarkable multilingual capabilities despite English-dominated pre-training, attributed to cross-lingual mechanisms during pre-training. Existing methods for enhancing cross-lingual transfer remain constrained by parallel resources, suffering from limited linguistic and domain coverage. We propose Cross-lingual In-context Pre-training (CrossIC-PT), a simple and scalable approach that enhances cross-lingual transfer by leveraging semantically related bilingual texts via simple next-word prediction. We construct CrossIC-PT samples by interleaving semantic-related bilingual Wikipedia documents into a single context window. To access window size constraints, we implement a systematic segmentation policy to split long bilingual document pairs into chunks while adjusting the sliding window mechanism to preserve contextual coherence. We further extend data availability through a semantic retrieval framework to construct CrossIC-PT samples from web-crawled corpus. Experimental results demonstrate that CrossIC-PT improves multilingual performance on three models (Llama-3.1-8B, Qwen2.5-7B, and Qwen2.5-1.5B) across six target languages, yielding performance gains of 3.79\%, 3.99\%, and 1.95\%, respectively, with additional improvements after data augmentation.
\end{abstract}

\section{Introduction}
Recent state-of-the-art (SOTA) large language models (LLMs)~\cite{Achiam2023GPT4TR,TheC3, Reid2024Gemini1U} have demonstrated remarkable multilingual capabilities. These models are typically pre-trained on massive web-crawled corpora, where English text overwhelmingly dominates in the quantity~\cite{Brown2020LanguageMA, Llama3}. However, current LLMs exhibit unexpectedly strong performance on non-English languages that cannot be fully explained by their relative data proportions during pre-training. Researchers have attributed this phenomenon to cross-lingual transfer in LLM training, where linguistic patterns and knowledge acquired from high-resource languages (particularly English) appear to transfer effectively to enhance performance on the other languages~\cite{XQuAD,BLOOM,Wang2024ProbingTE}.

\begin{figure}
    \centering
    \includegraphics[width=0.9\linewidth]{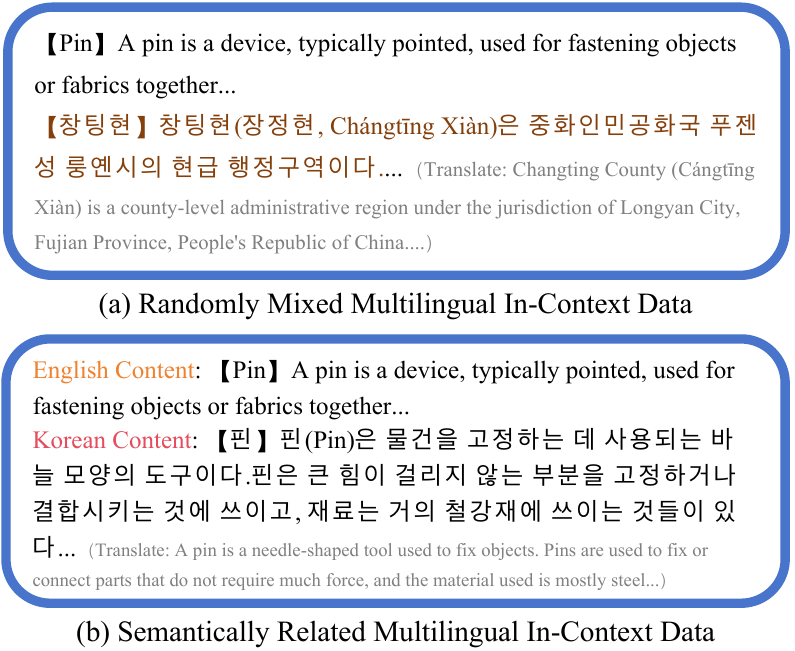}
    \caption{Existing works randomly mix multilingual texts (a) in an input window. Our approach groups semantically related texts (b) to enhance cross-lingual transfer.}
    \label{fig1}
\end{figure}

A series of works have explored methods for interpreting and enhancing cross-lingual transfer during language model pre-training. \citet{EMNLP22} revealed that even in English-dominated pre-training data, millions of non-English tokens can be identified, which are crucial for multilingual capabilities. Some studies have attempted to analyze cross-lingual transfer abilities from perspectives of shared vocabulary and representation similarity\cite{Patil2022OverlapbasedVG,Lin2023mPLMSimBC}, though their conclusions primarily apply to specific language groups. The predominant research paradigm has focused on explicitly enhancing cross-lingual transfer through exploiting supervision signals, such as parallel corpora\cite{Distillation,Marco-LLM,EMMA-500,Bilingual_Adaptation,parallel_data}, code-switching datasets\cite{Three_Pronged,code-switching}, or fine-grained signals like cross lingual entity links\cite{LEIA}. These approaches, however, remain constrained by the limited quantity, domain coverage, and morphological diversity of available bilingual resources (e.g., dictionaries, and parallel sentence pairs).

Our approach builds upon the fundamental principle of LLM pre-training: contextual modeling through next-word prediction (NWP) loss optimization within fixed-length text windows. Since LLMs could effectively learn monolingual semantics through this mechanism, we hypothesize that extending NWP optimization on semantically related cross-lingual content - using source language context to predict target language sequences - could enhance cross-lingual transfer capabilities. As illustrated in Fig.\ref{fig1}(b), our method constructs \textbf{Cross-lingual In-context} samples by interleaving semantically related bilingual text pairs. Subsequently, we optimize LLMs through standard NWP loss computation on these composite samples. The proposed \textbf{Cross}-lingual \textbf{I}n-\textbf{C}ontext \textbf{P}re-\textbf{T}raining (\textbf{CrossIC-PT}) eliminates the reliance on parallel corpora, and could be applied to different types of text, providing a simple and scalable paradigm for cross-lingual transfer learning.

\begin{figure*}
    \centering
    \includegraphics[width=0.9\linewidth]{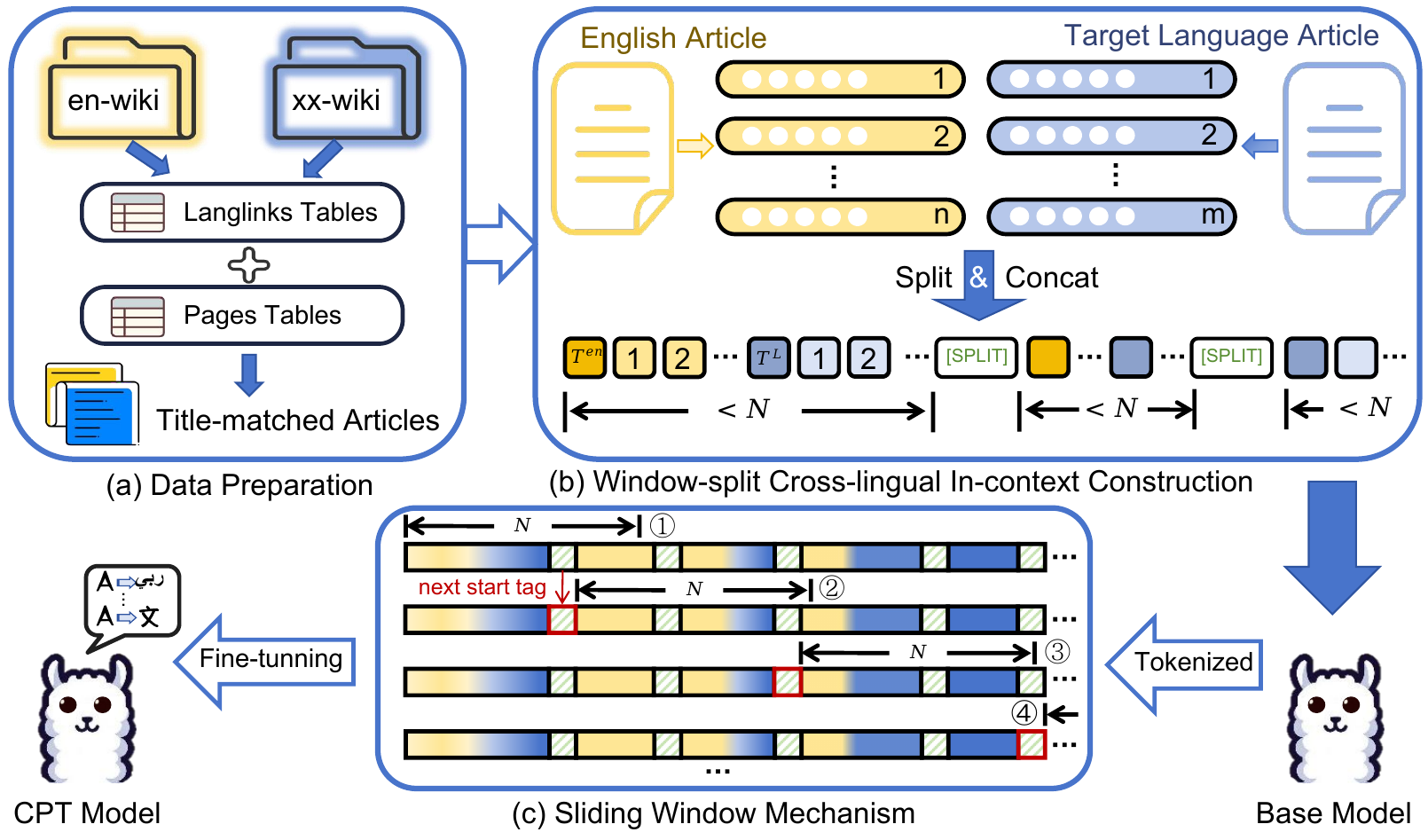}
    \caption{The implementation process of our method, CrossIC-PT, which constructs cross-lingual in-contexts based on Wikipedia data and performs continued pre-training (CPT) on existing multilingual models. Here, \(N\) represents the input window length of the model. The \(T\) indicates the title of the articles, and \(L\) indicates the target language.}
    \label{fig2}
\end{figure*}

To validate our method, we implement the proposed \textbf{CrossIC-PT} method through continued pre-training (CPT) on existing LLMs~\cite{Llama3, Qwen2.5}. This strategy converges faster than training from scratch, providing a cost-effective solution for multilingual experimentation~\cite{CPT_EMNLP}. Leveraging the readily available multilingual Wikipedia data, we construct a cross-lingual in-context corpus by concatenating two bilingual Wikipedia articles on the same entity, as illustrated in Fig.\ref{fig2}. To mitigate context window length constraints, we segment article pairs into bilingual sub-pairs, using a dedicated [SPLIT] token as delimiters (Fig.\ref{fig2}(b)). We further optimize the sliding window mechanism, ensuring that the next window starts from the token after the last [SPLIT] of the current window, thereby maintaining context coherence and enhancing cross-lingual alignment learning. To further assess the generalizability of our method, we develop a cross-lingual semantic retrieval framework build upon that extends beyond Wikipedia data by incorporating web-crawled text. As shown in Fig.\ref{fig3}, this framework retrieves semantically related paragraphs from the English Fineweb\_edu~\cite{fineweb-edu} dataset using title and partial content keywords from the target-language Wikipedia articles as query.

We conducted experiments in six languages based on three LLMs (Llama-3.1-8B, Qwen2.5-7B, Qwen2.5-1.5B) and tested them on seven tasks. The CrossIC-PT model, built on Wikipedia, improved average performance by 3.79\%, 3.99\%, and 1.95\% compared to the base models, respectively. The expansion of the data further boosted performance by 0.73\% for Llama-3.1-8B.

Our contributions can be summarized as follows:
\begin{itemize}[leftmargin=*,nolistsep]
\item We propose \textbf{CrossIC-PT}, a novel method that enhances LLMs' cross-lingual transfer by leveraging semantically related in-context data.
\item To address input window length limitations, we design a window-split strategy with a [SPLIT] token and an optimized sliding window mechanism to maintain cross-lingual contextual coherence.
\item We also design a cross-lingual semantic retrieval framework to augment training data, which further enhances model performance, proving the robustness and scalability of our approach.
\end{itemize}

\section{Related Work}


Many existing works focus on collecting multilingual data to enhance LLMs' cross-lingual capabilities~\cite{Qwen2.5,Llama3,Marco-LLM,EMMA-500}. Samples from different languages are randomly packed into fixed window sizes (e.g., 4096) without cross-contamination in self-attention. Even so, these models already demonstrate multilingual ability. Based on this, we hypothesize that concatenating semantically related English and target language data (Fig.\ref{fig1}(b)) could enhance cross-lingual transfer by leveraging implicit supervision signals.

Cross-lingual supervision signals have been proven effective in enhancing LLMs' cross-lingual transfer abilities~\cite{Three_Pronged,LEIA,PLUG}. Most methods rely on bilingual corpora as explicit supervision signals~\cite{Distillation,Marco-LLM,EMMA-500,Bilingual_Adaptation,parallel_data}. Some works, like~\cite{Distillation} distills translation pairs from LLMs through back-translation to create supervision signals. Others, such as~\cite{Three_Pronged,LEIA}, apply code-switching techniques to replace or augment words with English translations. ~\cite{code-switching} also explores code-switching at various levels using curriculum learning. However, parallel corpora have restricted types, domains (most bilingual corpora are short sentence level bitexts, and usually extracted from news websites), and quantity. Synthetic parallel documents built by back-translation, however, are limited in text Quality. In contrast, our method constructs semantically related document pairs from the authentic data on the Internet, which is more scalable and less problematic.

\section{Method}

Multilingual LLM pre-training typically packs documents from different languages randomly into the fixed-size context window. We hypothesize that concatenating semantically related English and target language corpora, predicting the next words based upon not only monolingual and cross-lingual context could enhance cross-lingual transfer ability. We call this concatenated sample \textbf{Cross-lingual In-context} data, where English serves as the guiding context for learning the target language. Based on this, we propose \textbf{CrossIC-PT}, a pre-training method leveraging cross-lingual in-context data.

As LLMs are pre-trained with a fixed tokens window size (e.g. 4096 tokens), cross-lingual in-context data, which are usually two times longer than the vanilla monolingual documents, may exceed the size limit. Simplifying the packing by length may break the cross-lingual relationship. To address this problem, we carefully design a bilingual-aware window-split strategy to construct cross-lingual in-context data. Additionally, to avoid the traditional sliding window mechanism from splitting the concatenated context, we further optimize the sliding window mechanism to ensure context coherence.

We take advantage of Wikipedia data to implement our method, as shown in Fig.\ref{fig2}, consisting of three key steps: (1) \textbf{Data preparation}, where we extract and align bilingual article pairs from Wikipedia (Sec.~\ref{sec31}); (2) \textbf{Window-split cross-lingual in-context construction}, where we split multilingual contexts to match the length of the input window (Sec.~\ref{sec32}); and (3) training with an optimized \textbf{sliding window mechanism} to enhance cross-lingual representation learning (Sec.~\ref{sec33}). In order to test the generalization of our approach, we propose a cross-lingual semantic retrieval framework to augment the training data (Sec.~\ref{sec34}).

\subsection{Data Preparation}  \label{sec31}
To obtain aligned article pairs in English and the target language (denoted \(L\)), we utilize three key tables from \textbf{Wikimedia} with three steps:  

1. \textbf{Langlinks Table for Language \(L\)}:  
   It contains article ID mappings between language \(L\) and other languages with matching titles, along with the corresponding title names \(T\). This table helps identify English article IDs and title names that match those in language \(L\), mapping as \((ID^L,(ID^{en},T^{en}))\).

2. \textbf{English Pages Table}:  
   The `pages` table of English provides article IDs and their corresponding title. We use it to remove English articles with blank or invalid titles from the initial mappings in step (1), yielding the final ID pairs \((ID^L,ID^{en})\).  

3. \textbf{Articles Tables for English and Language \(L\)}:  
   The `articles` tables for both languages contain the article ID and full information on the web page, which includes the article content. Using the bilingual article ID pairs \((ID^L,ID^{en})\), we extract the corresponding article pairs with matching titles. 
   
To ensure completeness, we also perform the reverse mapping \((ID^{en},ID^{L})\), and combine the results with the forward mappings to obtain a comprehensive set of bilingual article pairs. This process ensures that we capture all possible title-matched articles between English and the target language.

\subsection{Window-split Cross-lingual In-Context Construction} \label{sec32}

To fit within the context size \(N\), we set a strategy for processing long article pairs by segmenting them into paragraphs and aligning them sequentially. Specifically, for each bilingual article pair \((A_{en}, A_L)\), we extract the title \(T\) and split the articles into paragraphs by signal "\texttt{\textbackslash n\textbackslash n}":
\[
A_{en} = [p_1^{en}, p_2^{en}, ..., p_n^{en}], \quad A_L = [p_1^{L}, p_2^{L}, ..., p_m^{L}].
\]
We iteratively select paragraph pairs \((p_i^{en}, p_i^{L})\) until adding the \(k\)-th pair would exceed the length \(N\), and then concat the paragraphs as follows:
\[
(T^{en},p_1^{en}; p_2^{en}; ...; p_{k-1}^{en}; T^{L}, p_1^{L}; p_2^{L}; ...; p_{k-1}^{L}),
\]
with all English paragraphs preceding the target language \(L\) paragraphs, and the delimiter as "\texttt{\textbackslash n\textbackslash n}". Each concatenated sequence is terminated with a special \texttt{[SPLIT]} token to mark the end of the context window. If the paragraphs of one language are exhausted before the other, we continue concatenating paragraphs from the remaining language until the length limit \(N\) is reached or all paragraphs are used. This process converts each bilingual article pair into one or more window-split multilingual in-contexts, each fitting within the length limit \(N\).

\subsection{Pre-training Method} \label{sec33}
\subsubsection{Sliding Window Mechanism}
In standard pre-training, the sliding window mechanism concatenates all training data and sliding with a fixed window size. However, this can randomly break down our cross-lingual in-contexts, disrupting coherence. To address this, we optimize the sliding window by the introduced tag \texttt{"[SPLIT]"}. Specifically, all the windows set the start boundary after the last \texttt{"[SPLIT]"} token, as shown in Fig.~\ref{fig2}. The tokens remain between the end boundary and the latest \texttt{"[SPLIT]"} token will be dropped. In this way we could try best to preserve the cross-lingual coherence within the window.

\subsubsection{Training Strategy}

As discussed earlier, continual pre-training (CPT) is cost-effective for cross-lingual transfer. So we adopt it in all our experiments. Recent studies, such as \citep{Lora_cross}, show that Low-Rank Adaptation (LoRA) is highly competitive with full fine-tuning, especially in low-data and cross-lingual transfer scenarios. In our experiments,  we also adopt LoRA during continual pre-training and results reveal that LoRA consistently provides better and more stable performance.

\begin{figure}[t]
    \centering
    \includegraphics[width=0.85\linewidth]{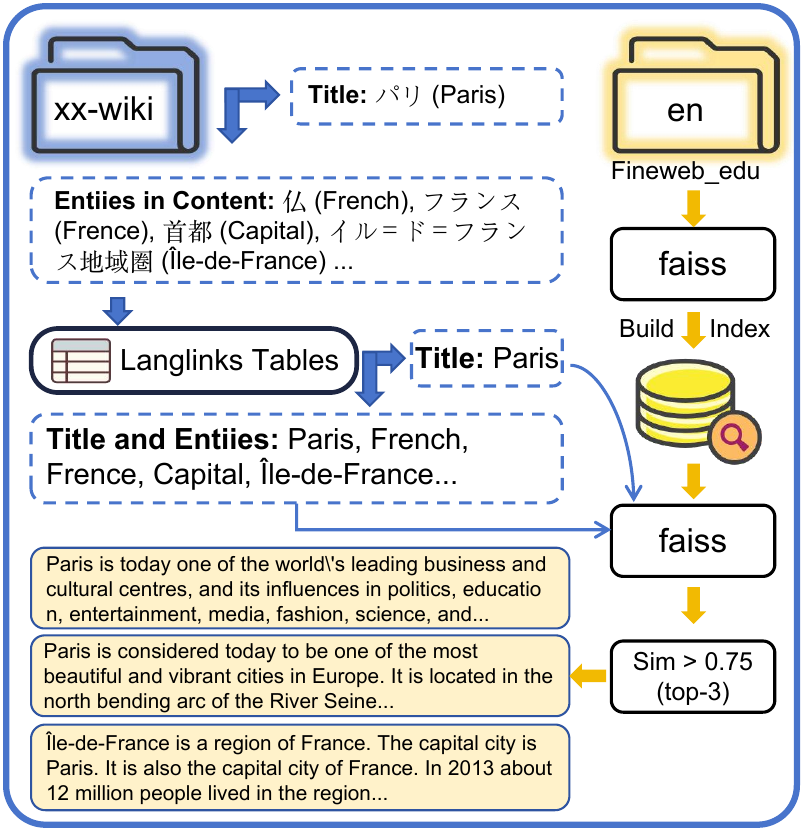}
    \caption{The framework of cross-lingual semantic retrieval based on FAISS similarity search tool.}
    \label{fig3}
\end{figure}

\subsection{Data Augmentation via Retrieval} \label{sec34}
To validate our approach, we use the Wikipedia corpus, which includes data in nearly 200 languages linked by matched titles. While the content across languages is not strictly parallel, it covers the same topics, making it suitable for our needs. To enhance the generalization of our method, we introduce a cross-lingual semantic retrieval framework based on the FAISS similarity search tool~\cite{FAISS}\footnote{We used the embedding from https://huggingface.co/Alibaba-NLP/gte-multilingual-base.}, as shown in Fig.\ref{fig3}. This framework augments the training data by incorporating relevant English articles from the Fineweb\_edu~\cite{fineweb-edu} dataset, retrieved using title and content keywords (up to 10 per article) extracted from the Wikipedia data.

First, keywords are extracted from the target-language Wikipedia page and mapped to English via the \texttt{langlinks} table. Fineweb\_edu is then indexed using FAISS for similarity calculations. We employ a two-step retrieval process using FAISS: (1) retrieval based on title keywords, and (2) retrieval based on both title and content keywords. The final similarity score is the average of these two steps, balancing the importance of the titles (which may be ambiguous) and content keywords. Based on empirical observations, we set a similarity threshold of 0.75 and retrieved up to three relevant samples per target-language article to construct window-split cross-lingual in-context data. These samples are combined with the original Wikipedia data to form an augmented dataset.

\section{Experiments}
\subsection{Training Data}
Our training data is primarily sourced from Wikipedia\footnote{We used the 20240720 wiki-dumps and processed them with wikiextractor (https://github.com/attardi/wikiextractor) to remove boilerplate text and extract article content.} (denoted as W), with token counts for English and each target language listed in Table~\ref{data_stats}. We selected six target languages \(L\): Arabic (ar), Spanish (es), Japanese (ja), Korean (ko), Portuguese (pt), and Thai (th). To further expand the dataset, we retrieved relevant English data from a subset of Fineweb\_edu (denoted as F), which has a file size of 17.44GB. The token counts for the augmented data are also provided in Table~\ref{data_stats}.

\begin{table}[htbp]
  \centering
    \resizebox{0.45\textwidth}{!}{\begin{tabular}{c|c|cccccc}
    \toprule
    data  & language & ar    & es    & ja    & ko    & pt    & th \\
    \midrule
    \multirow{2}[2]{*}{W} & en    & 1.53B & 1.88B & 1.32B & 1.01B & 1.48B & 0.42B \\
          & \(L\) & 0.67B & 1.57B & 1.28B & 0.37B & 0.81B & 0.18B \\
    \midrule
    \multirow{2}[2]{*}{F} & en    & 0.12B & 0.10B & 0.06B & 0.04B & 0.05B & 0.10B \\
          & \(L\) & 0.12B & 0.13B & 0.08B & 0.03B & 0.05B & 0.06B \\
    \bottomrule
    \end{tabular}}%
  \caption{The token counts for the data from Wikipedia (W) and augmented data from Wikipedia and Fineweb\_edu (F). }
  \label{data_stats}%
\end{table}%

\begin{table*}[htbp]
  \centering
    \resizebox{0.95\textwidth}{!}{\begin{tabular}{c|cc|cccccc|c}
    \toprule
    \multicolumn{3}{c}{\multirow{2}[4]{*}{Model}} & \multicolumn{6}{c}{Languages}       & \multicolumn{1}{c}{\multirow{2}[4]{*}{AVG}} \\
\cmidrule{4-9}    \multicolumn{3}{c}{} & \multicolumn{1}{c}{ar} & \multicolumn{1}{c}{es} & \multicolumn{1}{c}{ja} & \multicolumn{1}{c}{ko} & \multicolumn{1}{c}{pt} & \multicolumn{1}{c}{th} &  \\
    \midrule
    \multirow{9}[7]{*}{Llama-3.1-8B} & \multicolumn{2}{c}{base} & 37.96  & 42.11  & 43.02  & 43.82  & 44.36  & 38.79  & 41.68  \\
          & \multicolumn{2}{c}{LEIA} & \multicolumn{1}{c}{37.04±0.49} & \multicolumn{1}{c}{44.03±0.24} & \multicolumn{1}{c}{44.86±0.89} & \multicolumn{1}{c}{44.11±0.48} & \multicolumn{1}{c}{44.48±0.58} & \multicolumn{1}{c}{42.90±0.82} & \multicolumn{1}{c}{42.98±0.25} \\
          & \multicolumn{2}{c}{X-MONO-PT} & 39.11  & 44.35  & 43.57  & 44.24  & 44.04  & 42.09  & 42.90  \\
\cmidrule{2-10}          & \multicolumn{1}{c}{\multirow{2}[2]{*}{F}} & \multicolumn{1}{c}{Mix-PT} & 38.24  & 44.09  & 44.09  & 44.12  & 45.65  & 41.44  & 42.94  \\
          & \multicolumn{1}{c}{} & \multicolumn{1}{c}{\cellcolor[rgb]{ .851,  .882,  .957}CrossIC-PT} & \cellcolor[rgb]{ .851,  .882,  .957}39.66  & \cellcolor[rgb]{ .851,  .882,  .957}44.57  & \cellcolor[rgb]{ .851,  .882,  .957}45.61  & \cellcolor[rgb]{ .851,  .882,  .957}46.02  & \cellcolor[rgb]{ .851,  .882,  .957}47.32  & \cellcolor[rgb]{ .851,  .882,  .957}42.30  & \cellcolor[rgb]{ .851,  .882,  .957}44.24  \\
\cmidrule{2-10}          & \multicolumn{1}{c}{\multirow{2}[2]{*}{W}} & \multicolumn{1}{c}{Mix-PT} & 38.09  & 43.46  & 44.81  & 44.75  & 46.45  & 42.38  & 43.32  \\
          & \multicolumn{1}{c}{} & \multicolumn{1}{c}{\cellcolor[rgb]{ .851,  .882,  .957}CrossIC-PT} & \cellcolor[rgb]{ .851,  .882,  .957}40.57  & \cellcolor[rgb]{ .851,  .882,  .957}45.49  & \cellcolor[rgb]{ .851,  .882,  .957}47.27  & \cellcolor[rgb]{ .851,  .882,  .957}46.87  & \cellcolor[rgb]{ .851,  .882,  .957}49.09  & \cellcolor[rgb]{ .851,  .882,  .957}43.51  & \cellcolor[rgb]{ .851,  .882,  .957}45.47  \\
\cmidrule{2-10}          & \multicolumn{1}{c}{\multirow{2}[1]{*}{W+F}} & \multicolumn{1}{c}{Mix-PT} & 40.19  & 44.58  & 44.75  & 44.48  & 46.62  & 42.05  & 43.78  \\
          & \multicolumn{1}{c}{} & \multicolumn{1}{c}{\cellcolor[rgb]{ .851,  .882,  .957}CrossIC-PT} & \cellcolor[rgb]{ .851,  .882,  .957}\textbf{41.18 } & \cellcolor[rgb]{ .851,  .882,  .957}\textbf{46.93 } & \cellcolor[rgb]{ .851,  .882,  .957}\textbf{48.10 } & \cellcolor[rgb]{ .851,  .882,  .957}\textbf{47.32 } & \cellcolor[rgb]{ .851,  .882,  .957}\textbf{49.97 } & \cellcolor[rgb]{ .851,  .882,  .957}\textbf{43.72 } & \cellcolor[rgb]{ .851,  .882,  .957}\textbf{46.20 } \\
    \midrule
    \midrule
    \multirow{3}[0]{*}{Qwen-2.5-7B} & \multicolumn{2}{c}{base} & 50.91  & 54.71  & 56.95  & 55.52  & 56.49  & 53.81  & 54.73  \\
          & \multicolumn{2}{c}{Mix-PT} & 54.48  & 58.71  & 57.69  & 57.39  & 60.30  & 56.19  & 57.46  \\
          & \multicolumn{2}{c}{\cellcolor[rgb]{ .851,  .882,  .957}CrossIC-PT} & \cellcolor[rgb]{ .851,  .882,  .957}\textbf{55.97 } & \cellcolor[rgb]{ .851,  .882,  .957}\textbf{59.44 } & \cellcolor[rgb]{ .851,  .882,  .957}\textbf{59.00 } & \cellcolor[rgb]{ .851,  .882,  .957}\textbf{59.03 } & \cellcolor[rgb]{ .851,  .882,  .957}\textbf{61.59 } & \cellcolor[rgb]{ .851,  .882,  .957}\textbf{57.33 } & \cellcolor[rgb]{ .851,  .882,  .957}\textbf{58.73 } \\
    \midrule
    \midrule
    \multirow{3}[0]{*}{Qwen-2.5-1.5B} & \multicolumn{2}{c}{base} & 37.83  & 43.90  & 42.26  & 39.75  & 44.35  & 41.40  & 41.58  \\
          & \multicolumn{2}{c}{Mix-PT} & 38.14  & 44.37  & 41.85  & 39.48  & 45.63  & 40.92  & 41.73  \\
          & \multicolumn{2}{c}{\cellcolor[rgb]{ .851,  .882,  .957}CrossIC-PT} & \cellcolor[rgb]{ .851,  .882,  .957}\textbf{40.21 } & \cellcolor[rgb]{ .851,  .882,  .957}\textbf{45.09 } & \cellcolor[rgb]{ .851,  .882,  .957}\textbf{43.96 } & \cellcolor[rgb]{ .851,  .882,  .957}\textbf{41.47 } & \cellcolor[rgb]{ .851,  .882,  .957}\textbf{48.25 } & \cellcolor[rgb]{ .851,  .882,  .957}\textbf{42.23 } & \cellcolor[rgb]{ .851,  .882,  .957}\textbf{43.54 } \\\bottomrule
    \end{tabular}}%
  \caption{The average results of our CrossIC-PT model, based on three base LLMs (Llama-3.1-8B, Qwen2.5-7B, and Qwen2.5-1.5B), are compared with corresponding baselines across six target languages. The cross-lingual in-context datasets used in methods based on Qwen2.5-7B and Qwen2.5-1.5B are sourced from Wikipedia(W).}
  \label{base_results}%
\end{table*}%

\subsection{Training Settings}
We conducted experiments on three base models: Llama-3.1-8B~\cite{Llama3}, Qwen2.5-7B~\cite{Qwen2.5}, and Qwen2.5-1.5B~\cite{Qwen2.5}. For LoRA, we set the rank to 64, alpha to 128, and dropout to 0.05. The input window length \(N\) was set to 4096, with a batch size of 128. All models were trained for one epoch, using a warmup ratio of 0.05, a cosine learning rate scheduler, and the AdamW optimizer. We randomly selected 0.1\% of the data as the validation set, with a seed number of 32. For Llama-3.1-8B and Qwen2.5-7B, the models after one epoch of training were used as the final models. For Qwen2.5-1.5B, we validated the model every 100 steps and saved the checkpoint with the lowest validation loss as the final model. The training was performed on 8 A100 GPUs.

\subsection{Benchmark}

We evaluated our models on several tasks from the latest multilingual and multitask benchmark, P-MMEVAL~\cite{P_MMEval}, which includes: generation (FLORES-200~\cite{flores_200}), understanding (XNLI~\cite{xnli}, MHELLASWAG \footnote{https://huggingface.co/datasets/alexandrainst/m\_hellaswag}), knowledge (MMMLU\footnote{https://huggingface.co/datasets/openai/MMMLU}), logical reasoning (MLOGIQA), and mathematical reasoning (MGSM~\cite{MGSM}). To further assess the models' paragraph comprehension abilities, we incorporated a reading comprehension task (MRC). The MRC test data includes TydiQA-GoldP~\cite{TyDiQA} for Arabic (ar) and Korean (ko), XQuAD~\cite{XQuAD} for Spanish (es), Portuguese (pt), and Thai (th), and 1,200 samples from JaQuAD~\cite{JaQuAD} for Japanese (ja). Details of the evaluation setting can be found in Appendix~\ref{apx1}.

\subsection{Baselines}
In addition to the \textbf{base models} (Llama-3.1-8B, Qwen2.5-7B, and Qwen2.5-1.5B), we included the most relevant baseline \textbf{Mix-PT}, which uses title-matched article pairs (Fig.~\ref{fig2}(a)) for pre-training but without cross-lingual text concatenation. 

Based on Llama-3.1-8B we also included two other baselines:(1)\textbf{LEIA}~\cite{LEIA}: A method that randomly adds English translations of entities to target-language Wikipedia data for pre-training, leveraging cross-lingual entity supervision. We reproduced this method using the provided code to construct the data and perform CPT on Llama-3.1-8B, ensuring the target-language token count matched ours. We conducted experiments with three random seeds (32, 111, 222) and reported the mean and variance of the results. (2)\textbf{X-MONO-PT}: A method that uses the target language data from our title-matched article pairs in Fig.\ref{fig2}(a) for pre-training.

\subsection{Results}
\subsubsection{Base Results}
ALl average six languages results of the baselines and our method, based on different training data volume from Wikipedia (W) and Fineweb\_edu (F) are shown in Table~\ref{base_results}. Detailed results for each languages can be found in Appendix~\ref{apx2}. CrossIC-PT consistently improves the performance of the base LLMs and outperforms other baselines, demonstrating the effectiveness of using semantically related cross-lingual in-context corpora for pre-training.

\begin{figure*}
    \centering
    \includegraphics[width=0.95\linewidth]{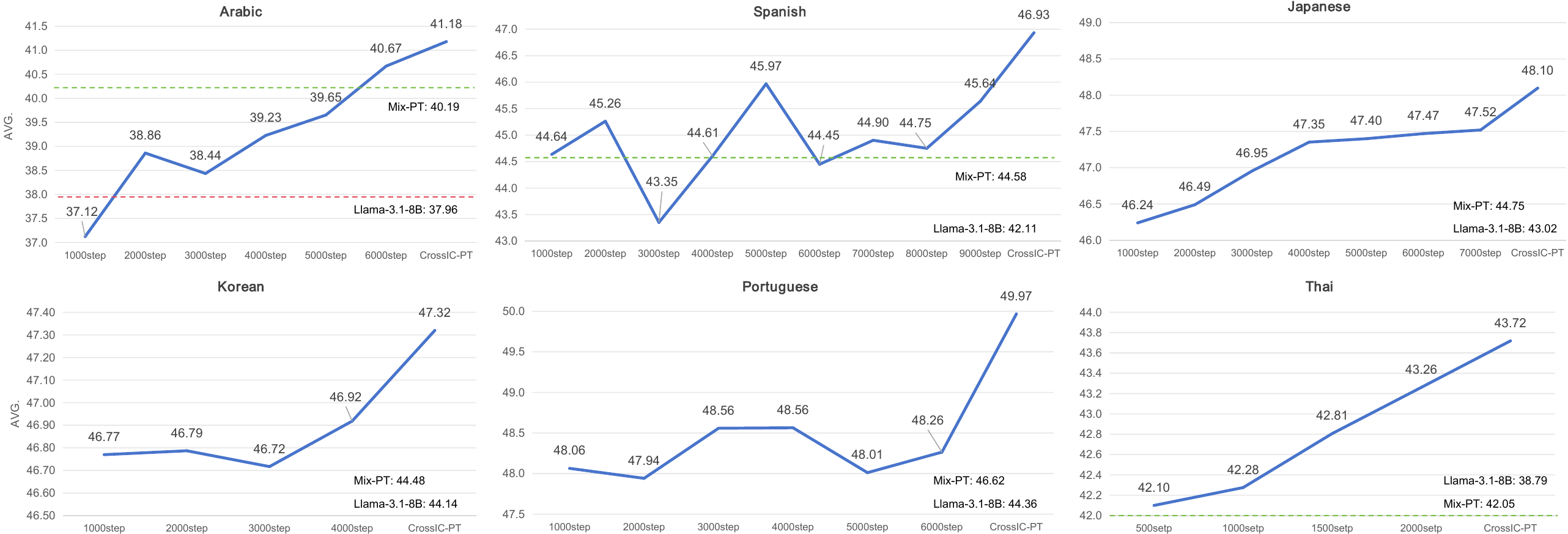}
    \caption{Performance progression of CrossIC-PT across intermediate checkpoints based on Llama-3.1-8B. Our method outperforms the baseline LLM early on, indicating quick acquisition of cross-lingual transfer capabilities, maintaining a slow upward trend as data volume increases.}
    \label{fig5}
\end{figure*}

Compared to the base LLMs, our CrossIC-PT method trained with only Wikipedia(W) data improves performance by 3.79\%, 3.99\%, and 1.95\% on Llama-3.1-8B, Qwen2.5-7B, and Qwen2.5-1.5B, respectively, across six languages. Notably, in Portuguese (pt), CrossIC-PT improves performance by 4.73\% on Llama-3.1-8B, surpassing the strongest baseline by 2.64\%. The performance gains for Qwen2.5 models are more pronounced as model size increases, which may be attributed to the fact that CPT performance is influenced by the initial capabilities of the model.


Our method consistently improves performance across all languages. The improvement in Thai is less noticeable on Qwen2.5-1.5B, likely due to the smaller dataset size. The LEIA method shows significant gains in some languages (Spanish, Japanese, and Thai), but its performance is unstable and data-dependent. For instance, the standard deviation for Japanese and Thai exceeds 0.8. This suggests that the implicit supervision signals from our cross-lingual in-context data are more robust and adaptable across languages compared to the entity-alignment signals used by LEIA.

The Mix-PT model is a strong baseline, trained on non-concatenated title-matched article pairs from Wikipedia, and improves performance across all six languages compared to the three base LLMs. However, our method improves the average performance by 2.15\% over the Mix-PT model on Llama-3.1-8B. Our method further enhances Mix-PT by concatenating cross-lingual data and designing an optimized sliding window mechanism.

\subsubsection{Results of Data Augmentation} 
To explore the generalization of our method, we propose a cross-lingual semantic retrieval framework (Fig.~\ref{fig3}) to augment the training data by expanding Wikipedia (W) with FineWeb-edu (F). After retrieval, the data volume increased by 0.06B–0.23B tokens. Although this is a relatively small increase, it improved the average performance of our method by 0.73\%. Even when using only the augmented data (F), our method still achieved a 1.26\% improvement over the strong baseline LEIA. This demonstrates that even if English data is not perfectly aligned with target languages, semantic similarity still facilitates cross-lingual transfer. Moreover, the simplicity of our retrieval process allows easy extension to various data sources.

Additionally, we saved several intermediate checkpoints to assess the impact of data volume on performance. As shown in Fig.~\ref{fig5}, at earlier checkpoints, our method outperformed the baseline LLM in all six languages and surpassed the strong baseline Mix-PT in four languages. This suggests that CrossIC-PT can quickly acquire useful cross-lingual transfer capabilities from the cross-lingual in-context data. Although performance improvements became slower as data volume increased, a consistent upward trend was still observed.

\begin{figure}
    \centering
    \includegraphics[width=0.95\linewidth]{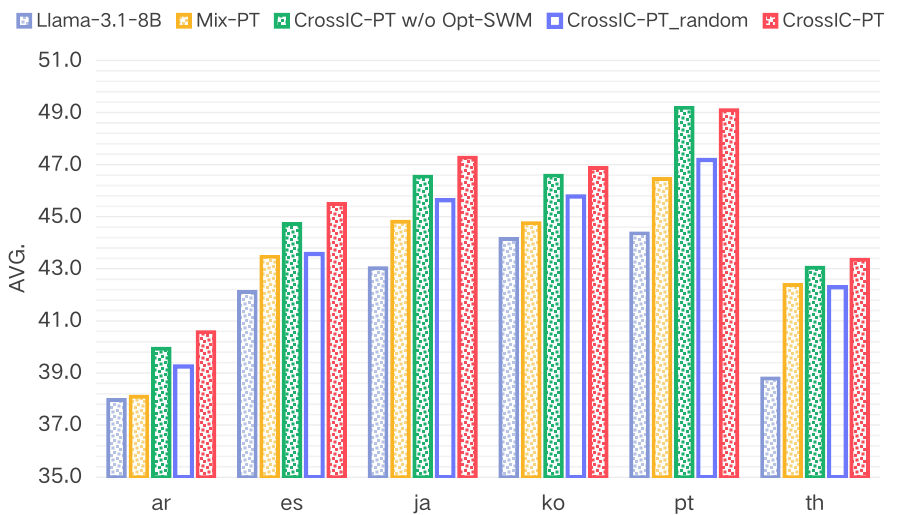}
    \caption{Ablation results of CrossIC-PT.}
    \label{fig4}
\end{figure}

\begin{table*}
  \centering
    \resizebox{0.94\textwidth}{!}{\tiny
    \begin{tabular}{l|cccccccc|c}
    \toprule
    \multicolumn{1}{c|}{\multirow{2}[3]{*}{Model}} & \multicolumn{1}{c}{\multirow{2}[3]{*}{XLOGIQA}} & \multicolumn{1}{c}{\multirow{2}[3]{*}{XHELLASWAG}} & \multicolumn{1}{c}{\multirow{2}[3]{*}{MMMLU}} & \multicolumn{1}{c}{\multirow{2}[3]{*}{XNLI}} & \multicolumn{1}{c}{\multirow{2}[3]{*}{MRC}} & \multicolumn{2}{c}{FLORESE-200} & \multicolumn{1}{c}{\multirow{2}[3]{*}{MGSM}} & \multicolumn{1}{|c}{\multirow{2}[3]{*}{AVG.}} \\
\cmidrule{7-8}    \multicolumn{1}{c|}{} &       &       &       &       &       & {en-xx} & xx-en &       &  \\
    \midrule
    Llama-3.1-8B & 33.25  & 35.33  & 40.10  & 56.17  & 57.49  & 38.56  & 29.41  & 38.00  & 41.04  \\
    Mix-PT & 34.75  & 36.68  & 43.05  & 59.17  & 60.36  & 39.63  & 32.72  & 36.96  & 42.92  \\
    CrossIC-PT & \textbf{36.00 } & \textbf{39.71 } & 43.15  & \textbf{62.17 } & \textbf{63.02}  & \textbf{41.39 } & 30.44  & \textbf{39.68 } & \textbf{44.44}  \\
    \midrule
    {CrossIC-PT\(_{mix}\)} & 34.75  & 32.33  & \textbf{43.55 } & 58.33  & {62.20 } & 40.75  & 33.53  & 36.96  & 42.80  \\
    {CrossIC-PT\(_{reverse}\)} & 35.50  & 33.69  & 43.00  & 57.67  & 62.41  & 39.51  & \textbf{34.12 } & 36.40  & 42.79  \\
    \bottomrule
    \end{tabular}}%
  \caption{The average task results of CrossIC-PT with mix two directions (CrossIC-PT\(_{mix}\)) and the reverse direction (CrossIC-PT\(_{reverse}\)) of cross-lingual in-context data.}
  \label{direction_results}%
\end{table*}%

\subsection{Ablation Study}
We conduct ablation studies to evaluate two key components of our approach: (1) the optimized sliding window mechanism (Opt-SWM) and (2) the semantic related of cross-lingual contexts. First, comparing CrossIC-PT with and without Opt-SWM (denoted as CrossIC-PT w/o Opt-SWM), Fig.\ref{fig4} shows that while window-split cross-lingual contexts alone improve performance across all languages, adding Opt-SWM further enhances results by maintaining context coherence. Second, when replacing semantically similar pairs with randomly paired bilingual documents (CrossIC-PT\_random), performance degrades significantly - though still surpassing Mix-PT baselines - confirming the importance of our semantic similarity strategy. These results collectively validate the effectiveness of each design component in CrossIC-PT.

\section{Analysis}

\subsection{Concatenation Direction}
We believe that placing semantically related English text before target-language content helps models better learn from cross-lingual contexts. Thus, we set the order as English first, followed by the target language. To verify if this direction is more beneficial, we analyze the concatenation order.

To evaluate the impact of concatenation direction on performance, we compare the original direction (English first, target language second) with the reverse direction (target language first, English second), as well as a 1:1 random mix of both directions. Previously, we only reported results for the en-xx direction in the translation task. In this experiment, we also provide results for the xx-en direction on FLORES-200.

The average results of six languages across tasks are presented in Table~\ref{direction_results}. The effect of data concatenation order on translation tasks is most pronounced and fits the intuition. The best translation performance occurs when the concatenation direction matches the translation direction. When combining both directions, CrossIC-PT consistently outperforms the Mix-PT method in translation tasks, showing that even non-parallel bilingual data improves translation. Overall, the English-first, target language-second concatenation gives the best results, aligning with our intention of using English as context to guide the target language learning.

\subsection{Performance on English Tasks}

\begin{figure}
    \centering
    \includegraphics[width=0.9\linewidth]{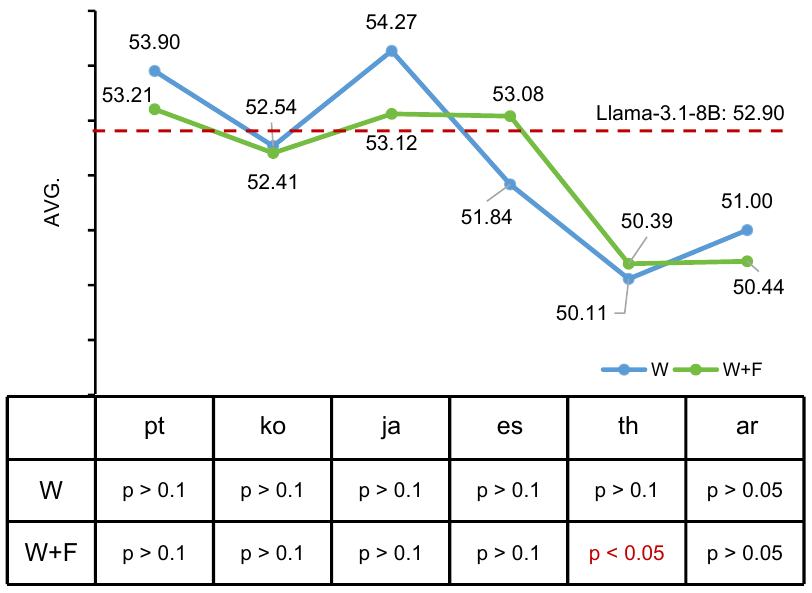}
    \caption{The average results of each target language model in English tasks. The \(p\) is the significant score between the CrossIC-PT model and Llama-3.1-8B.}
    \label{fig6}
\end{figure}
To prevent catastrophic forgetting during continual pre-training, it's important to ensure English performance is maintained. To verify this, we tested the performance of six target language models on English tasks, using the same tasks as before. The results are shown in Fig.\ref{fig6}.

The upper part of Fig.\ref{fig6} shows the average performance of each target language model on English tasks, with the x-axis ordered by the performance gap between Llama-3.1-8B’s performance on the target language and English. The trend suggests that a larger performance gap corresponds to a greater impact on English performance after training. For example, the English performance of Thai (th) and Arabic (ar) is lower. However, it is primarily due to a significant drop in one task. To further investigate, the lower part of Fig.\ref{fig6} presents the statistical significance ("p") of the performance differences between target language models and the base model, Llama-3.1-8B, across seven tasks. The results show that, except for the Thai model trained with data augmentation (which exhibits a significant drop in English performance), there are no significant differences for other target language models. This suggests that CrossIC-PT improves performance in target languages while effectively preserving English capabilities. We believe this is likely due to the inclusion of at least 50\% of English tokens in the cross-lingual in-context corpus, which helps mitigate severe forgetting. This result further validates the robustness and practicality of CrossIC-PT for cross-lingual transfer.

\section{Conclusion}
Our work explores a special angle by focusing on semantically related multilingual in-context to enhance the cross-lingual transfer capability of LLMs. We hypothesize that concatenating semantically related English and target language corpora as Cross-lingual In-context data is easily accessible and provides an implicitly cross-lingual supervision signal. Building on this hypothesis, we propose CrossIC-PT, a pre-training method based on cross-lingual in-context data. We implement our method using Wikipedia data and employ continual pre-training of existing LLMs on this data. To address the limitations posed by input window length during model training, we design a window-split strategy coupled with an optimized window sliding mechanism. Experimental results demonstrate that CrossIC-PT enhances multilingual performance across three models—Llama-3.1-8B, Qwen2.5-7B, and Qwen2.5-1.5B—across six target languages, achieving performance gains of 3.79\%, 3.99\%, and 1.95\%, respectively, compared to base models. Further improvements are observed after data augmentation using a semantic retrieval framework. Our approach is simple to scale for multilingual LLM pre-training and offers an efficient way to expand data volume.

\section*{Limitations}
To our knowledge, this work has the following limitations: 
\begin{itemize}
    \item Due to resource constraints, our experiments were limited to a context window length of 4096 tokens. Longer windows could better preserve the completeness of articles and enable the concatenation of similar multilingual data from more than two languages, potentially further enhancing cross-lingual transfer.
    \item Our experiments focused on validating the effectiveness of concatenated cross-lingual in-context data, so we performed continued pre-training on monolingual data rather than mixing multilingual data. While this choice aligns with our research goals, our approach also provides valuable insights for developers working on multilingual LLMs.
    \item Our data expansion method, based on retrieval, currently demonstrates how to retrieve additional English data from external sources using target-language Wikipedia data. However, this approach can be easily extended to retrieve more diverse data. Wikipedia's broad domain coverage makes it an ideal hub for retrieving both target language and English data from other sources. By controlling the retrieval process with appropriate similarity thresholds, the retrieved bilingual data can be used to construct high-quality cross-lingual in-context data.
\end{itemize}

\section*{Acknowledgements}
This work is supported by the National Natural Science Foundation of China (No. 62376245),  the Key Research and Development Program of Zhejiang Province, China (No. 2024C01034), the Fundamental Research Funds for the Central Universities  (226-2024-00170),  National Key Research and Development Project of China (No. 2018AAA0101900), and MOE Engineering Research Center of Digital Library, and the Alibaba Research Intern Program.
\bibliography{acl_latex}
\newpage
\appendix
\onecolumn
\section{The Setting for Evaluation} \label{apx1}
The prompts of each task we used are shown in Table \ref{task_prompts}. Since our method aims to transfer English capabilities to target languages, the prompts are primarily designed in English, and the demonstrations are also selected from English data. For the mathematical reasoning task (MGSM), we conducted an 8-shot test; for the reading comprehension task (MRC), we adopted a zero-shot setting to evaluate the model's understanding of the target language; for other tasks, we set up a 5-shot test. For multiple-choice tasks (e.g., XNLI, MMLU, XHELLASWAG, XLOGIQA), we directly obtain answers by predicting the next logits. For other tasks, we use greedy search to generate answers and extract the final answer through regular expression matching.

\begin{table*}[htbp]
\centering
\tiny
\begin{tabular}{p{0.2\textwidth}|p{0.7\textwidth}}
\toprule
\textbf{Task} & \textbf{Prompt} \\ \midrule
XLOGIQA & 
\texttt{Passage: \{context\}\textbackslash nQuestion: \{question\}\textbackslash nChoices:\textbackslash nA. \{option\_a\}\textbackslash nB. \{option\_b\}\textbackslash nC. \{option\_c\}\textbackslash nD. \{option\_d\}\textbackslash nAnswer:} \\ \midrule

XHELLASWAG & 
\texttt{\{premise\}\textbackslash nOptions: \textbackslash nA. \{option\_1\}\textbackslash nB. \{option\_2\}\textbackslash nC. \{option\_3\}\textbackslash nD. \{option\_4\}\textbackslash nQuestion: Which is the correct ending for the sentence from A, B, C, and D? \textbackslash nAnswer:} \\ \midrule

MMMLU & 
\texttt{The following is a multiple-choice question.\textbackslash n\textbackslash n\{question\}\textbackslash nA. \{option\_a\}\textbackslash nB. \{option\_b\}\textbackslash nC. \{option\_c\}\textbackslash nD. \{option\_d\}\textbackslash n\textbackslash nAnswer:} \\ \midrule

XNLI & 
\texttt{Take the following as truth: \{premise\}\textbackslash nThen the following statement: "\{hypothesis\}" is\textbackslash nOptions:\textbackslash nA. true\textbackslash nB. inconclusive\textbackslash nC. false\textbackslash nAnswer:} \\ \midrule

MRC & 
\texttt{Refer to the passage below and answer the following question:\textbackslash nPassage: \{context\}\textbackslash nQuestion: \{question\}\textbackslash nAnswer: Based on the passage, the answer to the question is "} \\ \midrule

FLORES-200 & 
\texttt{Translate from [source] to [target].\textbackslash n[source]: </X>\textbackslash n[target]:} \\ \midrule

MGSM & 
\texttt{Solve this math problem. Give the reasoning steps before giving the final answer on the last line by itself in the format of "[The answer is ]". Do not add anything other than the integer answer after "The answer is".\textbackslash n\textbackslash n\{question\}} \\ \bottomrule
\end{tabular}
\caption{Task Prompts."[]" represents optional content. For the FLORESE task, the "[source]" indicates the source language, and "[target]" indicates the target language of translation. For MGSM, "[The answer is ]" is the translation of "The answer is " according to the test language.}
\label{task_prompts}
\end{table*}

For the FLORES-200 generation tasks, we employed SacreBLEU\footnote{\url{https://github.com/mjpost/sacreBLEU}} with the default configuration: \textit{nrefs:1|case:mixed|eff:no|tok:13a|smooth:exp|version:2.0.0}. For languages lacking whitespace-based word boundaries (Japanese [ja], Korean [ko], and Thai [th]), we introduced a pre-tokenization step prior to BLEU computation, implemented as follows:

{
\small

\begin{lstlisting}
class NonASCIITokenizer(object):
    def __init__(self):
        self.is_cjk = re.compile(
            "([\u2e80-\u9fff]|"  # zh,ja,ko
            "[\ua960-\ua97f]|"  # Hangul Extended A
            "[\uac00-\ud7ff]|"  # Hangeul Syllables + Hangeul Letters Extension B
            "[\u0E00-\u0E7F]"  # th
            ")"
        )

    def __call__(self, sent):
        sent = sent.strip()
        chs = list(sent)
        line_chtok = []
        for ch in chs:
            if self.is_cjk.match(ch):
                line_chtok.append(' ')
                line_chtok.append(ch)
                line_chtok.append(' ')
            else:
                line_chtok.append(ch)
        line_chtok = trim_multiple_space(line_chtok)
        return ' '.join(line_chtok)

\end{lstlisting}
}

\section{Results of Per-Languages} \label{apx2}
The results of our method, ablation study, and the baselines across six languages in each task are shown in Table~\ref{task_results}.

\begin{table}[htbp]
  \centering
    \resizebox{0.98\textwidth}{!}{\begin{tabular}{c|c|c|ccccccc|c}
    \toprule
    \multicolumn{3}{c}{\multirow{2}[2]{*}{\textbf{Model}}} & \multicolumn{7}{c}{\textbf{Arabic Tasks}}   & \multicolumn{1}{c}{\multirow{2}[2]{*}{\textbf{AVG}}} \\
    \multicolumn{3}{c}{} & \multicolumn{1}{c}{\textbf{XLOGIQA}} & \multicolumn{1}{c}{\textbf{XHELLASWAG}} & \multicolumn{1}{c}{\textbf{MMMLU}} & \multicolumn{1}{c}{\textbf{XNLI}} & \multicolumn{1}{c}{\textbf{MRC}} & \multicolumn{1}{c}{\textbf{FLORES-200}} & \multicolumn{1}{c}{\textbf{MGSM}} &  \\
    \midrule
    \multirow{11}[7]{*}{Llama-3.1-8B} & \multicolumn{2}{c}{base} & 37.50  & 35.34  & 33.25  & 54.17  & 52.84  & 16.20  & 34.40  & 37.67  \\
          & \multicolumn{2}{c}{LEIA} & 33.75±1.77 & 32.76±0.70 & 34.33±0.24 & 51.39±1.04 & 53.02±1.56 & 18.55±0.17 & 35.47±0.75 & 37.04±0.49 \\
          & \multicolumn{2}{c}{X-MONO-PT} & 40.00  & 29.31  & 36.75  & 54.17  & 61.24  & 17.91  & 34.40  & 39.11  \\
\cmidrule{2-11}          & \multicolumn{1}{c}{\multirow{2}[2]{*}{F}} & \multicolumn{1}{c}{Mix-PT} & 33.75  & 33.62  & 33.75  & 57.50  & 58.41  & 17.04  & 33.60  & 38.24  \\
          & \multicolumn{1}{c}{} & \multicolumn{1}{c}{\cellcolor[rgb]{ .851,  .882,  .957}CrossIC-PT} & \cellcolor[rgb]{ .851,  .882,  .957}40.00  & \cellcolor[rgb]{ .851,  .882,  .957}32.76  & \cellcolor[rgb]{ .851,  .882,  .957}33.25  & \cellcolor[rgb]{ .851,  .882,  .957}60.83  & \cellcolor[rgb]{ .851,  .882,  .957}59.50  & \cellcolor[rgb]{ .851,  .882,  .957}17.26  & \cellcolor[rgb]{ .851,  .882,  .957}34.00  & \cellcolor[rgb]{ .851,  .882,  .957}39.66  \\
\cmidrule{2-11}          & \multicolumn{1}{c}{\multirow{4}[2]{*}{W}} & \multicolumn{1}{c}{Mix-PT} & 36.25  & 29.31  & 36.50  & 50.00  & 63.43  & 17.92  & 33.20  & 38.09  \\
          & \multicolumn{1}{c}{} & \multicolumn{1}{c}{\cellcolor[rgb]{ .851,  .882,  .957}CrossIC-PT w/o Opt-SWM} & \cellcolor[rgb]{ .851,  .882,  .957}33.75  & \cellcolor[rgb]{ .851,  .882,  .957}34.48  & \cellcolor[rgb]{ .851,  .882,  .957}36.00  & \cellcolor[rgb]{ .851,  .882,  .957}55.83  & \cellcolor[rgb]{ .851,  .882,  .957}63.65  & \cellcolor[rgb]{ .851,  .882,  .957}21.03  & \cellcolor[rgb]{ .851,  .882,  .957}34.80  & \cellcolor[rgb]{ .851,  .882,  .957}39.93  \\
          & \multicolumn{1}{c}{} & \multicolumn{1}{c}{\cellcolor[rgb]{ .851,  .882,  .957}CrossIC-PT\_random} & \cellcolor[rgb]{ .851,  .882,  .957}37.50  & \cellcolor[rgb]{ .851,  .882,  .957}33.62  & \cellcolor[rgb]{ .851,  .882,  .957}36.75  & \cellcolor[rgb]{ .851,  .882,  .957}51.67  & \cellcolor[rgb]{ .851,  .882,  .957}60.34  & \cellcolor[rgb]{ .851,  .882,  .957}18.04  & \cellcolor[rgb]{ .851,  .882,  .957}36.80  & \cellcolor[rgb]{ .851,  .882,  .957}39.25  \\
          & \multicolumn{1}{c}{} & \multicolumn{1}{c}{\cellcolor[rgb]{ .851,  .882,  .957}CrossIC-PT} & \cellcolor[rgb]{ .851,  .882,  .957}35.00  & \cellcolor[rgb]{ .851,  .882,  .957}35.34  & \cellcolor[rgb]{ .851,  .882,  .957}37.25  & \cellcolor[rgb]{ .851,  .882,  .957}55.83  & \cellcolor[rgb]{ .851,  .882,  .957}62.83  & \cellcolor[rgb]{ .851,  .882,  .957}22.95  & \cellcolor[rgb]{ .851,  .882,  .957}34.80  & \cellcolor[rgb]{ .851,  .882,  .957}40.57  \\
\cmidrule{2-11}          & \multicolumn{1}{c}{\multirow{2}[1]{*}{W+F}} & \multicolumn{1}{c}{Mix-PT} & 33.75  & 33.62  & 36.50  & 55.83  & 68.56  & 17.85  & 35.20  & 40.19  \\
          & \multicolumn{1}{c}{} & \multicolumn{1}{c}{\cellcolor[rgb]{ .851,  .882,  .957}CrossIC-PT} & \cellcolor[rgb]{ .851,  .882,  .957}36.25  & \cellcolor[rgb]{ .851,  .882,  .957}35.34  & \cellcolor[rgb]{ .851,  .882,  .957}37.25  & \cellcolor[rgb]{ .851,  .882,  .957}54.17  & \cellcolor[rgb]{ .851,  .882,  .957}66.81  & \cellcolor[rgb]{ .851,  .882,  .957}22.82  & \cellcolor[rgb]{ .851,  .882,  .957}35.60  & \cellcolor[rgb]{ .851,  .882,  .957}41.18  \\
    \midrule
    \midrule
    \multirow{3}[0]{*}{Qwen-2.5-7B} & \multicolumn{2}{c}{base} & 50.00  & 54.31  & 45.50  & 57.50  & 67.25  & 16.61  & 65.20  & 50.91  \\
          & \multicolumn{2}{c}{Cross-CPT} & 47.50  & 57.76  & 45.25  & 76.67  & 73.47  & 17.09  & 63.60  & 54.48  \\
          & \multicolumn{2}{c}{\cellcolor[rgb]{ .851,  .882,  .957}CrossIC-CPT} & \cellcolor[rgb]{ .851,  .882,  .957}50.00  & \cellcolor[rgb]{ .851,  .882,  .957}61.21  & \cellcolor[rgb]{ .851,  .882,  .957}46.00  & \cellcolor[rgb]{ .851,  .882,  .957}71.67  & \cellcolor[rgb]{ .851,  .882,  .957}73.58  & \cellcolor[rgb]{ .851,  .882,  .957}24.15  & \cellcolor[rgb]{ .851,  .882,  .957}65.20  & \cellcolor[rgb]{ .851,  .882,  .957}55.97  \\
    \midrule
    \midrule
    \multirow{3}[1]{*}{Qwen-2.5-1.5B} & \multicolumn{2}{c}{base} & 36.25  & 31.03  & 36.00  & 45.83  & 73.25  & 6.87  & 35.60  & 37.83  \\
          & \multicolumn{2}{c}{Cross-CPT} & 40.00  & 30.17  & 38.25  & 49.17  & 72.71  & 5.89  & 30.80  & 38.14  \\
          & \multicolumn{2}{c}{\cellcolor[rgb]{ .851,  .882,  .957}CrossIC-CPT} & \cellcolor[rgb]{ .851,  .882,  .957}41.25  & \cellcolor[rgb]{ .851,  .882,  .957}35.34  & \cellcolor[rgb]{ .851,  .882,  .957}37.25  & \cellcolor[rgb]{ .851,  .882,  .957}53.33  & \cellcolor[rgb]{ .851,  .882,  .957}76.09  & \cellcolor[rgb]{ .851,  .882,  .957}7.44  & \cellcolor[rgb]{ .851,  .882,  .957}30.80  & \cellcolor[rgb]{ .851,  .882,  .957}40.21  \\
    \midrule
    \multicolumn{1}{r}{} & \multicolumn{1}{r}{} & \multicolumn{1}{r}{} &       &       &       &       &       &       & \multicolumn{1}{r}{} &  \\
    \midrule
    \multicolumn{3}{c}{\multirow{2}[2]{*}{\textbf{Model}}} & \multicolumn{7}{c}{\textbf{Spanish Tasks}}  & \multicolumn{1}{c}{\multirow{2}[2]{*}{\textbf{AVG}}} \\
    \multicolumn{3}{c}{} & \multicolumn{1}{c}{\textbf{XLOGIQA}} & \multicolumn{1}{c}{\textbf{XHELLASWAG}} & \multicolumn{1}{c}{\textbf{MMMLU}} & \multicolumn{1}{c}{\textbf{XNLI}} & \multicolumn{1}{c}{\textbf{MRC}} & \multicolumn{1}{c}{\textbf{FLORES-200}} & \multicolumn{1}{c}{\textbf{MGSM}} &  \\
    \midrule
    \multirow{11}[7]{*}{Llama-3.1-8B} & \multicolumn{2}{c}{base} & 36.25  & 36.97  & 41.25  & 60.00  & 54.51  & 25.86  & 43.20  & 42.58  \\
          & \multicolumn{2}{c}{LEIA} & 38.75±1.77 & 43.70±0.00 & 39.08±0.51 & 63.06±2.83 & 52.69±0.34 & 25.60±0.09 & 45.33±0.82 & 44.03±0.24 \\
          & \multicolumn{2}{c}{X-MONO-PT} & 42.50  & 33.61  & 44.00  & 63.33  & 54.60  & 26.04  & 46.40  & 44.35  \\
\cmidrule{2-11}          & \multicolumn{1}{c}{\multirow{2}[2]{*}{F}} & \multicolumn{1}{c}{Mix-PT} & 41.50  & 37.82  & 43.25  & 62.50  & 51.48  & 25.69  & 46.40  & 44.09  \\
          & \multicolumn{1}{c}{} & \multicolumn{1}{c}{\cellcolor[rgb]{ .851,  .882,  .957}CrossIC-PT} & \cellcolor[rgb]{ .851,  .882,  .957}42.50  & \cellcolor[rgb]{ .851,  .882,  .957}37.82  & \cellcolor[rgb]{ .851,  .882,  .957}43.25  & \cellcolor[rgb]{ .851,  .882,  .957}63.33  & \cellcolor[rgb]{ .851,  .882,  .957}52.49  & \cellcolor[rgb]{ .851,  .882,  .957}25.80  & \cellcolor[rgb]{ .851,  .882,  .957}46.80  & \cellcolor[rgb]{ .851,  .882,  .957}44.57  \\
\cmidrule{2-11}          & \multicolumn{1}{c}{\multirow{4}[2]{*}{W}} & \multicolumn{1}{c}{Mix-PT} & 37.50  & 39.50  & 44.25  & 57.50  & 56.71  & 25.98  & 42.80  & 43.46  \\
          & \multicolumn{1}{c}{} & \multicolumn{1}{c}{\cellcolor[rgb]{ .851,  .882,  .957}CrossIC-PT w/o Opt-SWM} & \cellcolor[rgb]{ .851,  .882,  .957}40.00  & \cellcolor[rgb]{ .851,  .882,  .957}42.86  & \cellcolor[rgb]{ .851,  .882,  .957}43.25  & \cellcolor[rgb]{ .851,  .882,  .957}57.50  & \cellcolor[rgb]{ .851,  .882,  .957}59.32  & \cellcolor[rgb]{ .851,  .882,  .957}27.33  & \cellcolor[rgb]{ .851,  .882,  .957}42.80  & \cellcolor[rgb]{ .851,  .882,  .957}44.72  \\
          & \multicolumn{1}{c}{} & \multicolumn{1}{c}{\cellcolor[rgb]{ .851,  .882,  .957}CrossIC-PT\_random} & \cellcolor[rgb]{ .851,  .882,  .957}43.75  & \cellcolor[rgb]{ .851,  .882,  .957}35.29  & \cellcolor[rgb]{ .851,  .882,  .957}44.00  & \cellcolor[rgb]{ .851,  .882,  .957}55.00  & \cellcolor[rgb]{ .851,  .882,  .957}56.96  & \cellcolor[rgb]{ .851,  .882,  .957}25.99  & \cellcolor[rgb]{ .851,  .882,  .957}44.00  & \cellcolor[rgb]{ .851,  .882,  .957}43.57  \\
          & \multicolumn{1}{c}{} & \multicolumn{1}{c}{\cellcolor[rgb]{ .851,  .882,  .957}CrossIC-PT} & \cellcolor[rgb]{ .851,  .882,  .957}36.25  & \cellcolor[rgb]{ .851,  .882,  .957}43.70  & \cellcolor[rgb]{ .851,  .882,  .957}44.75  & \cellcolor[rgb]{ .851,  .882,  .957}60.83  & \cellcolor[rgb]{ .851,  .882,  .957}58.65  & \cellcolor[rgb]{ .851,  .882,  .957}27.82  & \cellcolor[rgb]{ .851,  .882,  .957}46.40  & \cellcolor[rgb]{ .851,  .882,  .957}45.49  \\
\cmidrule{2-11}          & \multicolumn{1}{c}{\multirow{2}[1]{*}{W+F}} & \multicolumn{1}{c}{Mix-PT} & 36.25  & 42.86  & 45.00  & 59.17  & 56.71  & 25.68  & 46.40  & 44.58  \\
          & \multicolumn{1}{c}{} & \multicolumn{1}{c}{\cellcolor[rgb]{ .851,  .882,  .957}CrossIC-PT} & \cellcolor[rgb]{ .851,  .882,  .957}38.75  & \cellcolor[rgb]{ .851,  .882,  .957}47.90  & \cellcolor[rgb]{ .851,  .882,  .957}43.50  & \cellcolor[rgb]{ .851,  .882,  .957}62.50  & \cellcolor[rgb]{ .851,  .882,  .957}63.71  & \cellcolor[rgb]{ .851,  .882,  .957}27.77  & \cellcolor[rgb]{ .851,  .882,  .957}44.40  & \cellcolor[rgb]{ .851,  .882,  .957}46.93  \\
    \midrule
    \midrule
    \multirow{3}[0]{*}{Qwen-2.5-7B} & \multicolumn{2}{c}{base} & 42.50  & 65.55  & 51.50  & 66.67  & 55.02  & 25.35  & 76.40  & 54.71  \\
          & \multicolumn{2}{c}{Cross-CPT} & 47.50  & 63.87  & 52.75  & 80.00  & 66.16  & 25.11  & 75.60  & 58.71  \\
          & \multicolumn{2}{c}{\cellcolor[rgb]{ .851,  .882,  .957}CrossIC-CPT} & \cellcolor[rgb]{ .851,  .882,  .957}42.50  & \cellcolor[rgb]{ .851,  .882,  .957}68.07  & \cellcolor[rgb]{ .851,  .882,  .957}51.50  & \cellcolor[rgb]{ .851,  .882,  .957}82.50  & \cellcolor[rgb]{ .851,  .882,  .957}69.45  & \cellcolor[rgb]{ .851,  .882,  .957}27.66  & \cellcolor[rgb]{ .851,  .882,  .957}74.40  & \cellcolor[rgb]{ .851,  .882,  .957}59.44  \\
    \midrule
    \midrule
    \multirow{3}[1]{*}{Qwen-2.5-1.5B} & \multicolumn{2}{c}{base} & 38.75  & 46.22  & 48.50  & 50.83  & 51.31  & 20.06  & 51.60  & 43.90  \\
          & \multicolumn{2}{c}{Cross-CPT} & 35.00  & 42.02  & 47.00  & 56.67  & 61.18  & 19.49  & 49.20  & 44.37  \\
          & \multicolumn{2}{c}{\cellcolor[rgb]{ .851,  .882,  .957}CrossIC-CPT} & \cellcolor[rgb]{ .851,  .882,  .957}40.00  & \cellcolor[rgb]{ .851,  .882,  .957}43.70  & \cellcolor[rgb]{ .851,  .882,  .957}47.50  & \cellcolor[rgb]{ .851,  .882,  .957}55.83  & \cellcolor[rgb]{ .851,  .882,  .957}56.46  & \cellcolor[rgb]{ .851,  .882,  .957}22.12  & \cellcolor[rgb]{ .851,  .882,  .957}50.00  & \cellcolor[rgb]{ .851,  .882,  .957}45.09  \\
    \midrule
    \multicolumn{1}{r}{} & \multicolumn{1}{r}{} & \multicolumn{1}{r}{} &       &       &       &       &       &       & \multicolumn{1}{r}{} &  \\
    \midrule
    \multicolumn{3}{c}{\multirow{2}[2]{*}{\textbf{Model}}} & \multicolumn{7}{c}{\textbf{Japanese Tasks}} & \multicolumn{1}{c}{\multirow{2}[2]{*}{\textbf{AVG}}} \\
    \multicolumn{3}{c}{} & \multicolumn{1}{c}{\textbf{XLOGIQA}} & \multicolumn{1}{c}{\textbf{XHELLASWAG}} & \multicolumn{1}{c}{\textbf{MMMLU}} & \multicolumn{1}{c}{\textbf{XNLI}} & \multicolumn{1}{c}{\textbf{MRC}} & \multicolumn{1}{c}{\textbf{FLORES-200}} & \multicolumn{1}{c}{\textbf{MGSM}} &  \\
    \midrule
    \multirow{11}[7]{*}{Llama-3.1-8B} & \multicolumn{2}{c}{base} & 32.50  & 35.00  & 36.50  & 54.17  & 68.95  & 39.81  & 33.20  & 42.88  \\
          & \multicolumn{2}{c}{LEIA} & 36.08±1.56 & 38.78±2.58 & 37.83±1.23 & 60.56±2.75 & 71.67±0.51 & 39.22±0.16 & 29.87±0.19 & 44.86±0.89 \\
          & \multicolumn{2}{c}{X-MONO-PT} & 32.50  & 33.33  & 39.50  & 54.17  & 73.25  & 40.64  & 31.60  & 43.57  \\
\cmidrule{2-11}          & \multicolumn{1}{c}{\multirow{2}[2]{*}{F}} & \multicolumn{1}{c}{Mix-PT} & 32.50  & 36.67  & 38.75  & 55.00  & 70.50  & 40.04  & 35.20  & 44.09  \\
          & \multicolumn{1}{c}{} & \multicolumn{1}{c}{\cellcolor[rgb]{ .851,  .882,  .957}CrossIC-PT} & \cellcolor[rgb]{ .851,  .882,  .957}40.00  & \cellcolor[rgb]{ .851,  .882,  .957}35.83  & \cellcolor[rgb]{ .851,  .882,  .957}39.75  & \cellcolor[rgb]{ .851,  .882,  .957}57.50  & \cellcolor[rgb]{ .851,  .882,  .957}70.83  & \cellcolor[rgb]{ .851,  .882,  .957}40.13  & \cellcolor[rgb]{ .851,  .882,  .957}35.20  & \cellcolor[rgb]{ .851,  .882,  .957}45.61  \\
\cmidrule{2-11}          & \multicolumn{1}{c}{\multirow{4}[2]{*}{W}} & \multicolumn{1}{c}{Mix-PT} & 33.75  & 37.50  & 41.00  & 55.83  & 72.57  & 40.63  & 32.40  & 44.81  \\
          & \multicolumn{1}{c}{} & \multicolumn{1}{c}{\cellcolor[rgb]{ .851,  .882,  .957}CrossIC-PT w/o Opt-SWM} & \cellcolor[rgb]{ .851,  .882,  .957}35.00  & \cellcolor[rgb]{ .851,  .882,  .957}40.00  & \cellcolor[rgb]{ .851,  .882,  .957}39.25  & \cellcolor[rgb]{ .851,  .882,  .957}59.17  & \cellcolor[rgb]{ .851,  .882,  .957}75.37  & \cellcolor[rgb]{ .851,  .882,  .957}41.76  & \cellcolor[rgb]{ .851,  .882,  .957}35.20  & \cellcolor[rgb]{ .851,  .882,  .957}46.54  \\
          & \multicolumn{1}{c}{} & \multicolumn{1}{c}{\cellcolor[rgb]{ .851,  .882,  .957}CrossIC-PT\_random} & \cellcolor[rgb]{ .851,  .882,  .957}40.00  & \cellcolor[rgb]{ .851,  .882,  .957}32.50  & \cellcolor[rgb]{ .851,  .882,  .957}40.00  & \cellcolor[rgb]{ .851,  .882,  .957}57.50  & \cellcolor[rgb]{ .851,  .882,  .957}75.17  & \cellcolor[rgb]{ .851,  .882,  .957}40.65  & \cellcolor[rgb]{ .851,  .882,  .957}33.60  & \cellcolor[rgb]{ .851,  .882,  .957}45.63  \\
          & \multicolumn{1}{c}{} & \multicolumn{1}{c}{\cellcolor[rgb]{ .851,  .882,  .957}CrossIC-PT} & \cellcolor[rgb]{ .851,  .882,  .957}35.00  & \cellcolor[rgb]{ .851,  .882,  .957}40.00  & \cellcolor[rgb]{ .851,  .882,  .957}39.25  & \cellcolor[rgb]{ .851,  .882,  .957}61.67  & \cellcolor[rgb]{ .851,  .882,  .957}77.08  & \cellcolor[rgb]{ .851,  .882,  .957}42.29  & \cellcolor[rgb]{ .851,  .882,  .957}35.60  & \cellcolor[rgb]{ .851,  .882,  .957}47.27  \\
\cmidrule{2-11}          & \multicolumn{1}{c}{\multirow{2}[1]{*}{W+F}} & \multicolumn{1}{c}{Mix-PT} & 31.25  & 40.00  & 42.00  & 51.67  & 75.92  & 40.78  & 31.60  & 44.75  \\
          & \multicolumn{1}{c}{} & \multicolumn{1}{c}{\cellcolor[rgb]{ .851,  .882,  .957}CrossIC-PT} & \cellcolor[rgb]{ .851,  .882,  .957}33.75  & \cellcolor[rgb]{ .851,  .882,  .957}47.50  & \cellcolor[rgb]{ .851,  .882,  .957}39.75  & \cellcolor[rgb]{ .851,  .882,  .957}65.00  & \cellcolor[rgb]{ .851,  .882,  .957}75.58  & \cellcolor[rgb]{ .851,  .882,  .957}41.90  & \cellcolor[rgb]{ .851,  .882,  .957}33.20  & \cellcolor[rgb]{ .851,  .882,  .957}48.10  \\
    \midrule
    \midrule
    \multirow{3}[0]{*}{Qwen-2.5-7B} & \multicolumn{2}{c}{base} & 48.75  & 58.33  & 49.50  & 65.00  & 75.75  & 40.91  & 60.40  & 56.95  \\
          & \multicolumn{2}{c}{Cross-CPT} & 47.50  & 60.83  & 51.75  & 75.83  & 71.00  & 40.11  & 56.80  & 57.69  \\
          & \multicolumn{2}{c}{\cellcolor[rgb]{ .851,  .882,  .957}CrossIC-CPT} & \cellcolor[rgb]{ .851,  .882,  .957}50.00  & \cellcolor[rgb]{ .851,  .882,  .957}58.33  & \cellcolor[rgb]{ .851,  .882,  .957}53.25  & \cellcolor[rgb]{ .851,  .882,  .957}76.67  & \cellcolor[rgb]{ .851,  .882,  .957}75.37  & \cellcolor[rgb]{ .851,  .882,  .957}42.98  & \cellcolor[rgb]{ .851,  .882,  .957}56.40  & \cellcolor[rgb]{ .851,  .882,  .957}59.00  \\
    \midrule
    \midrule
    \multirow{3}[1]{*}{Qwen-2.5-1.5B} & \multicolumn{2}{c}{base} & 38.75  & 33.33  & 41.50  & 52.50  & 69.00  & 27.91  & 32.80  & 42.26  \\
          & \multicolumn{2}{c}{Cross-CPT} & 38.75  & 35.83  & 42.25  & 55.83  & 69.83  & 23.65  & 26.80  & 41.85  \\
          & \multicolumn{2}{c}{\cellcolor[rgb]{ .851,  .882,  .957}CrossIC-CPT} & \cellcolor[rgb]{ .851,  .882,  .957}45.00  & \cellcolor[rgb]{ .851,  .882,  .957}38.33  & \cellcolor[rgb]{ .851,  .882,  .957}41.50  & \cellcolor[rgb]{ .851,  .882,  .957}57.50  & \cellcolor[rgb]{ .851,  .882,  .957}70.75  & \cellcolor[rgb]{ .851,  .882,  .957}26.24  & \cellcolor[rgb]{ .851,  .882,  .957}28.40  & \cellcolor[rgb]{ .851,  .882,  .957}43.96  \\
    \bottomrule
    \end{tabular}}%
    \caption{The results of our method, ablation study, and the baselines in Arabic, Spanish and Japanese.}
  \label{task_results}%
\end{table}%

\begin{table}[htbp]
  \centering
    \resizebox{\textwidth}{!}{\begin{tabular}{c|c|c|ccccccc|c}
    \toprule
    \multicolumn{3}{c}{\multirow{2}[2]{*}{\textbf{Model}}} & \multicolumn{7}{c}{\textbf{Korean Tasks}}   & \multicolumn{1}{c}{\multirow{2}[2]{*}{\textbf{AVG}}} \\
    \multicolumn{3}{c}{} & \multicolumn{1}{c}{\textbf{XLOGIQA}} & \multicolumn{1}{c}{\textbf{XHELLASWAG}} & \multicolumn{1}{c}{\textbf{MMMLU}} & \multicolumn{1}{c}{\textbf{XNLI}} & \multicolumn{1}{c}{\textbf{MRC}} & \multicolumn{1}{c}{\textbf{FLORES-200}} & \multicolumn{1}{c}{\textbf{MGSM}} &  \\
    \midrule
    \multirow{11}[7]{*}{Llama-3.1-8B} & \multicolumn{2}{c}{base} & 32.50  & 32.50  & 41.50  & 60.00  & 74.91  & 33.60  & 34.00  & 44.14  \\
          & \multicolumn{2}{c}{LEIA} & 36.00±2.70 & 32.28±1.04 & 41.03±0.24 & 63.83±0.68 & 74.31±0.63 & 28.98±0.34 & 35.60±0.65 & 44.58±0.46 \\
          & \multicolumn{2}{c}{X-MONO-PT} & 35.00  & 28.33  & 42.25  & 63.33  & 74.17  & 33.40  & 33.20  & 44.24  \\
\cmidrule{2-11}          & \multicolumn{1}{c}{\multirow{2}[2]{*}{F}} & \multicolumn{1}{c}{Mix-PT} & 36.25  & 28.33  & 39.75  & 60.83  & 75.65  & 33.20  & 34.80  & 44.12  \\
          & \multicolumn{1}{c}{} & \multicolumn{1}{c}{\cellcolor[rgb]{ .851,  .882,  .957}CrossIC-PT} & \cellcolor[rgb]{ .851,  .882,  .957}38.75  & \cellcolor[rgb]{ .851,  .882,  .957}32.50  & \cellcolor[rgb]{ .851,  .882,  .957}40.25  & \cellcolor[rgb]{ .851,  .882,  .957}65.00  & \cellcolor[rgb]{ .851,  .882,  .957}75.65  & \cellcolor[rgb]{ .851,  .882,  .957}34.00  & \cellcolor[rgb]{ .851,  .882,  .957}36.00  & \cellcolor[rgb]{ .851,  .882,  .957}46.02  \\
\cmidrule{2-11}          & \multicolumn{1}{c}{\multirow{4}[2]{*}{W}} & \multicolumn{1}{c}{Mix-PT} & 36.25  & 29.17  & 43.00  & 65.00  & 75.65  & 33.40  & 30.80  & 44.75  \\
          & \multicolumn{1}{c}{} & \multicolumn{1}{c}{\cellcolor[rgb]{ .851,  .882,  .957}CrossIC-PT w/o Opt-SWM} & \cellcolor[rgb]{ .851,  .882,  .957}37.50  & \cellcolor[rgb]{ .851,  .882,  .957}34.17  & \cellcolor[rgb]{ .851,  .882,  .957}41.75  & \cellcolor[rgb]{ .851,  .882,  .957}67.50  & \cellcolor[rgb]{ .851,  .882,  .957}77.12  & \cellcolor[rgb]{ .851,  .882,  .957}34.80  & \cellcolor[rgb]{ .851,  .882,  .957}33.20  & \cellcolor[rgb]{ .851,  .882,  .957}46.58  \\
          & \multicolumn{1}{c}{} & \multicolumn{1}{c}{\cellcolor[rgb]{ .851,  .882,  .957}CrossIC-PT\_random} & \cellcolor[rgb]{ .851,  .882,  .957}38.75  & \cellcolor[rgb]{ .851,  .882,  .957}32.50  & \cellcolor[rgb]{ .851,  .882,  .957}42.00  & \cellcolor[rgb]{ .851,  .882,  .957}65.00  & \cellcolor[rgb]{ .851,  .882,  .957}76.38  & \cellcolor[rgb]{ .851,  .882,  .957}34.26  & \cellcolor[rgb]{ .851,  .882,  .957}31.60  & \cellcolor[rgb]{ .851,  .882,  .957}45.78  \\
          & \multicolumn{1}{c}{} & \multicolumn{1}{c}{\cellcolor[rgb]{ .851,  .882,  .957}CrossIC-PT} & \cellcolor[rgb]{ .851,  .882,  .957}38.75  & \cellcolor[rgb]{ .851,  .882,  .957}33.33  & \cellcolor[rgb]{ .851,  .882,  .957}43.50  & \cellcolor[rgb]{ .851,  .882,  .957}66.67  & \cellcolor[rgb]{ .851,  .882,  .957}76.94  & \cellcolor[rgb]{ .851,  .882,  .957}34.89  & \cellcolor[rgb]{ .851,  .882,  .957}34.00  & \cellcolor[rgb]{ .851,  .882,  .957}46.87  \\
\cmidrule{2-11}          & \multicolumn{1}{c}{\multirow{2}[1]{*}{W+F}} & \multicolumn{1}{c}{Mix-PT} & 36.25  & 27.50  & 42.50  & 65.00  & 76.38  & 33.70  & 30.00  & 44.48  \\
          & \multicolumn{1}{c}{} & \multicolumn{1}{c}{\cellcolor[rgb]{ .851,  .882,  .957}CrossIC-PT} & \cellcolor[rgb]{ .851,  .882,  .957}40.00  & \cellcolor[rgb]{ .851,  .882,  .957}34.17  & \cellcolor[rgb]{ .851,  .882,  .957}43.00  & \cellcolor[rgb]{ .851,  .882,  .957}67.50  & \cellcolor[rgb]{ .851,  .882,  .957}77.12  & \cellcolor[rgb]{ .851,  .882,  .957}35.05  & \cellcolor[rgb]{ .851,  .882,  .957}34.40  & \cellcolor[rgb]{ .851,  .882,  .957}47.32  \\
    \midrule
    \midrule
    \multirow{3}[0]{*}{Qwen-2.5-7B} & \multicolumn{2}{c}{base} & 48.75  & 59.17  & 48.50  & 61.67  & 79.34  & 30.42  & 60.80  & 55.52  \\
          & \multicolumn{2}{c}{Cross-CPT} & 46.25  & 62.50  & 50.00  & 73.33  & 81.55  & 29.67  & 58.40  & 57.39  \\
          & \multicolumn{2}{c}{\cellcolor[rgb]{ .851,  .882,  .957}CrossIC-CPT} & \cellcolor[rgb]{ .851,  .882,  .957}51.25  & \cellcolor[rgb]{ .851,  .882,  .957}62.50  & \cellcolor[rgb]{ .851,  .882,  .957}49.00  & \cellcolor[rgb]{ .851,  .882,  .957}75.00  & \cellcolor[rgb]{ .851,  .882,  .957}81.55  & \cellcolor[rgb]{ .851,  .882,  .957}35.51  & \cellcolor[rgb]{ .851,  .882,  .957}58.40  & \cellcolor[rgb]{ .851,  .882,  .957}59.03  \\
    \midrule
    \midrule
    \multirow{3}[1]{*}{Qwen-2.5-1.5B} & \multicolumn{2}{c}{base} & 32.50  & 31.67  & 40.00  & 54.17  & 70.48  & 15.84  & 33.60  & 39.75  \\
          & \multicolumn{2}{c}{Cross-CPT} & 35.00  & 27.50  & 39.75  & 63.33  & 72.32  & 12.83  & 25.60  & 39.48  \\
          & \multicolumn{2}{c}{\cellcolor[rgb]{ .851,  .882,  .957}CrossIC-CPT} & \cellcolor[rgb]{ .851,  .882,  .957}35.00  & \cellcolor[rgb]{ .851,  .882,  .957}31.67  & \cellcolor[rgb]{ .851,  .882,  .957}41.00  & \cellcolor[rgb]{ .851,  .882,  .957}66.67  & \cellcolor[rgb]{ .851,  .882,  .957}72.32  & \cellcolor[rgb]{ .851,  .882,  .957}16.40  & \cellcolor[rgb]{ .851,  .882,  .957}27.20  & \cellcolor[rgb]{ .851,  .882,  .957}41.47  \\
    \midrule
    \multicolumn{1}{r}{} & \multicolumn{1}{r}{} & \multicolumn{1}{r}{} &       &       &       &       &       &       & \multicolumn{1}{r}{} &  \\
    \midrule
    \multicolumn{3}{c}{\multirow{2}[2]{*}{\textbf{Model}}} & \multicolumn{7}{c}{\textbf{Portugues Tasks}} & \multicolumn{1}{c}{\multirow{2}[2]{*}{\textbf{AVG}}} \\
    \multicolumn{3}{c}{} & \multicolumn{1}{c}{\textbf{XLOGIQA}} & \multicolumn{1}{c}{\textbf{XHELLASWAG}} & \multicolumn{1}{c}{\textbf{MMMLU}} & \multicolumn{1}{c}{\textbf{XNLI}} & \multicolumn{1}{c}{\textbf{MRC}} & \multicolumn{1}{c}{\textbf{FLORES-200}} & \multicolumn{1}{c}{\textbf{MGSM}} &  \\
    \midrule
    \multirow{11}[7]{*}{Llama-3.1-8B} & \multicolumn{2}{c}{base} & 33.75  & 40.52  & 42.75  & 56.67  & 49.42  & 44.37  & 46.80  & 44.90  \\
          & \multicolumn{2}{c}{LEIA} & 32.50±1.77 & 42.09±2.67 & 40.67±0.51 & 61.61±1.42 & 45.47±1.03 & 44.35±0.08 & 44.67±1.47 & 44.48±0.58 \\
          & \multicolumn{2}{c}{X-MONO-PT} & 35.00  & 36.21  & 45.25  & 53.33  & 50.25  & 45.07  & 43.20  & 44.04  \\
\cmidrule{2-11}          & \multicolumn{1}{c}{\multirow{2}[2]{*}{F}} & \multicolumn{1}{c}{Mix-PT} & 40.00  & 37.07  & 41.75  & 60.00  & 49.08  & 44.47  & 47.20  & 45.65  \\
          & \multicolumn{1}{c}{} & \multicolumn{1}{c}{\cellcolor[rgb]{ .851,  .882,  .957}CrossIC-PT} & \cellcolor[rgb]{ .851,  .882,  .957}40.00  & \cellcolor[rgb]{ .851,  .882,  .957}37.93  & \cellcolor[rgb]{ .851,  .882,  .957}43.25  & \cellcolor[rgb]{ .851,  .882,  .957}60.00  & \cellcolor[rgb]{ .851,  .882,  .957}59.83  & \cellcolor[rgb]{ .851,  .882,  .957}44.62  & \cellcolor[rgb]{ .851,  .882,  .957}45.60  & \cellcolor[rgb]{ .851,  .882,  .957}47.32  \\
\cmidrule{2-11}          & \multicolumn{1}{c}{\multirow{4}[2]{*}{W}} & \multicolumn{1}{c}{Mix-PT} & 35.00  & 41.38  & 46.50  & 59.17  & 52.67  & 45.26  & 45.20  & 46.45  \\
          & \multicolumn{1}{c}{} & \multicolumn{1}{c}{\cellcolor[rgb]{ .851,  .882,  .957}CrossIC-PT w/o Opt-SWM} & \cellcolor[rgb]{ .851,  .882,  .957}38.75  & \cellcolor[rgb]{ .851,  .882,  .957}46.55  & \cellcolor[rgb]{ .851,  .882,  .957}45.75  & \cellcolor[rgb]{ .851,  .882,  .957}65.00  & \cellcolor[rgb]{ .851,  .882,  .957}53.67  & \cellcolor[rgb]{ .851,  .882,  .957}47.33  & \cellcolor[rgb]{ .851,  .882,  .957}47.20  & \cellcolor[rgb]{ .851,  .882,  .957}49.18  \\
          & \multicolumn{1}{c}{} & \multicolumn{1}{c}{\cellcolor[rgb]{ .851,  .882,  .957}CrossIC-PT\_random} & \cellcolor[rgb]{ .851,  .882,  .957}38.75  & \cellcolor[rgb]{ .851,  .882,  .957}37.93  & \cellcolor[rgb]{ .851,  .882,  .957}44.75  & \cellcolor[rgb]{ .851,  .882,  .957}62.50  & \cellcolor[rgb]{ .851,  .882,  .957}57.67  & \cellcolor[rgb]{ .851,  .882,  .957}45.06  & \cellcolor[rgb]{ .851,  .882,  .957}43.60  & \cellcolor[rgb]{ .851,  .882,  .957}47.18  \\
          & \multicolumn{1}{c}{} & \multicolumn{1}{c}{\cellcolor[rgb]{ .851,  .882,  .957}CrossIC-PT} & \cellcolor[rgb]{ .851,  .882,  .957}37.50  & \cellcolor[rgb]{ .851,  .882,  .957}44.83  & \cellcolor[rgb]{ .851,  .882,  .957}47.00  & \cellcolor[rgb]{ .851,  .882,  .957}62.50  & \cellcolor[rgb]{ .851,  .882,  .957}56.67  & \cellcolor[rgb]{ .851,  .882,  .957}47.94  & \cellcolor[rgb]{ .851,  .882,  .957}47.20  & \cellcolor[rgb]{ .851,  .882,  .957}49.09  \\
\cmidrule{2-11}          & \multicolumn{1}{c}{\multirow{2}[1]{*}{W+F}} & \multicolumn{1}{c}{Mix-PT} & 36.25  & 39.66  & 44.00  & 60.00  & 53.92  & 44.84  & 43.60  & 46.04  \\
          & \multicolumn{1}{c}{} & \multicolumn{1}{c}{\cellcolor[rgb]{ .851,  .882,  .957}CrossIC-PT} & \cellcolor[rgb]{ .851,  .882,  .957}37.50  & \cellcolor[rgb]{ .851,  .882,  .957}44.83  & \cellcolor[rgb]{ .851,  .882,  .957}47.50  & \cellcolor[rgb]{ .851,  .882,  .957}65.83  & \cellcolor[rgb]{ .851,  .882,  .957}56.33  & \cellcolor[rgb]{ .851,  .882,  .957}47.81  & \cellcolor[rgb]{ .851,  .882,  .957}46.80  & \cellcolor[rgb]{ .851,  .882,  .957}49.51  \\
    \midrule
    \midrule
    \multirow{3}[0]{*}{Qwen-2.5-7B} & \multicolumn{2}{c}{base} & 43.75  & 65.52  & 50.50  & 65.00  & 52.92  & 43.36  & 74.40  & 56.49  \\
          & \multicolumn{2}{c}{Cross-CPT} & 47.50  & 64.66  & 52.00  & 80.00  & 59.75  & 42.96  & 75.20  & 60.30  \\
          & \multicolumn{2}{c}{\cellcolor[rgb]{ .851,  .882,  .957}CrossIC-CPT} & \cellcolor[rgb]{ .851,  .882,  .957}50.00  & \cellcolor[rgb]{ .851,  .882,  .957}67.24  & \cellcolor[rgb]{ .851,  .882,  .957}52.00  & \cellcolor[rgb]{ .851,  .882,  .957}80.83  & \cellcolor[rgb]{ .851,  .882,  .957}60.33  & \cellcolor[rgb]{ .851,  .882,  .957}48.33  & \cellcolor[rgb]{ .851,  .882,  .957}72.40  & \cellcolor[rgb]{ .851,  .882,  .957}61.59  \\
    \midrule
    \midrule
    \multirow{3}[1]{*}{Qwen-2.5-1.5B} & \multicolumn{2}{c}{base} & 40.00  & 42.24  & 46.25  & 48.33  & 49.75  & 32.70  & 51.20  & 44.35  \\
          & \multicolumn{2}{c}{Cross-CPT} & 38.75  & 41.38  & 47.50  & 55.83  & 55.25  & 30.73  & 50.00  & 45.63  \\
          & \multicolumn{2}{c}{\cellcolor[rgb]{ .851,  .882,  .957}CrossIC-CPT} & \cellcolor[rgb]{ .851,  .882,  .957}36.25  & \cellcolor[rgb]{ .851,  .882,  .957}45.69  & \cellcolor[rgb]{ .851,  .882,  .957}48.75  & \cellcolor[rgb]{ .851,  .882,  .957}62.50  & \cellcolor[rgb]{ .851,  .882,  .957}55.67  & \cellcolor[rgb]{ .851,  .882,  .957}38.86  & \cellcolor[rgb]{ .851,  .882,  .957}50.00  & \cellcolor[rgb]{ .851,  .882,  .957}48.25  \\
    \midrule
    \multicolumn{1}{r}{} & \multicolumn{1}{r}{} & \multicolumn{1}{r}{} &       &       &       &       &       &       & \multicolumn{1}{r}{} &  \\
    \midrule
    \multicolumn{3}{c}{\multirow{2}[2]{*}{\textbf{Model}}} & \multicolumn{7}{c}{\textbf{Thai Tasks}}     & \multicolumn{1}{c}{\multirow{2}[2]{*}{\textbf{AVG}}} \\
    \multicolumn{3}{c}{} & \multicolumn{1}{c}{\textbf{XLOGIQA}} & \multicolumn{1}{c}{\textbf{XHELLASWAG}} & \multicolumn{1}{c}{\textbf{MMMLU}} & \multicolumn{1}{c}{\textbf{XNLI}} & \multicolumn{1}{c}{\textbf{MRC}} & \multicolumn{1}{c}{\textbf{FLORES-200}} & \multicolumn{1}{c}{\textbf{MGSM}} &  \\
    \midrule
    \multirow{11}[7]{*}{Llama-3.1-8B} & \multicolumn{2}{c}{base} & 31.25  & 31.67  & 38.50  & 50.00  & 39.66  & 49.14  & 32.80  & 39.00  \\
          & \multicolumn{2}{c}{LEIA} & 32.50±1.02 & 36.83±2.04 & 37.92±0.51 & 59.56±1.57 & 50.95±0.39 & 49.37±0.36 & 33.20±2.04 & 42.90±0.82 \\
          & \multicolumn{2}{c}{X-MONO-PT} & 30.00  & 34.17  & 40.75  & 57.50  & 47.43  & 52.76  & 32.00  & 42.09  \\
\cmidrule{2-11}          & \multicolumn{1}{c}{\multirow{2}[2]{*}{F}} & \multicolumn{1}{c}{Mix-PT} & 31.25  & 29.17  & 39.50  & 51.67  & 51.39  & 51.90  & 35.20  & 41.44  \\
          & \multicolumn{1}{c}{} & \multicolumn{1}{c}{\cellcolor[rgb]{ .851,  .882,  .957}CrossIC-PT} & \cellcolor[rgb]{ .851,  .882,  .957}31.25  & \cellcolor[rgb]{ .851,  .882,  .957}30.00  & \cellcolor[rgb]{ .851,  .882,  .957}39.50  & \cellcolor[rgb]{ .851,  .882,  .957}55.00  & \cellcolor[rgb]{ .851,  .882,  .957}52.67  & \cellcolor[rgb]{ .851,  .882,  .957}51.65  & \cellcolor[rgb]{ .851,  .882,  .957}36.00  & \cellcolor[rgb]{ .851,  .882,  .957}42.30  \\
\cmidrule{2-11}          & \multicolumn{1}{c}{\multirow{4}[2]{*}{W}} & \multicolumn{1}{c}{Mix-PT} & 31.25  & 35.83  & 40.50  & 58.33  & 44.22  & 52.90  & 33.60  & 42.38  \\
          & \multicolumn{1}{c}{} & \multicolumn{1}{c}{\cellcolor[rgb]{ .851,  .882,  .957}CrossIC-PT w/o Opt-SWM} & \cellcolor[rgb]{ .851,  .882,  .957}32.50  & \cellcolor[rgb]{ .851,  .882,  .957}36.67  & \cellcolor[rgb]{ .851,  .882,  .957}40.25  & \cellcolor[rgb]{ .851,  .882,  .957}59.17  & \cellcolor[rgb]{ .851,  .882,  .957}45.32  & \cellcolor[rgb]{ .851,  .882,  .957}53.80  & \cellcolor[rgb]{ .851,  .882,  .957}33.60  & \cellcolor[rgb]{ .851,  .882,  .957}43.04  \\
          & \multicolumn{1}{c}{} & \multicolumn{1}{c}{\cellcolor[rgb]{ .851,  .882,  .957}CrossIC-PT\_random} & \cellcolor[rgb]{ .851,  .882,  .957}35.00  & \cellcolor[rgb]{ .851,  .882,  .957}35.83  & \cellcolor[rgb]{ .851,  .882,  .957}42.00  & \cellcolor[rgb]{ .851,  .882,  .957}54.17  & \cellcolor[rgb]{ .851,  .882,  .957}43.29  & \cellcolor[rgb]{ .851,  .882,  .957}52.92  & \cellcolor[rgb]{ .851,  .882,  .957}32.80  & \cellcolor[rgb]{ .851,  .882,  .957}42.29  \\
          & \multicolumn{1}{c}{} & \multicolumn{1}{c}{\cellcolor[rgb]{ .851,  .882,  .957}CrossIC-PT} & \cellcolor[rgb]{ .851,  .882,  .957}32.50  & \cellcolor[rgb]{ .851,  .882,  .957}36.67  & \cellcolor[rgb]{ .851,  .882,  .957}41.25  & \cellcolor[rgb]{ .851,  .882,  .957}59.17  & \cellcolor[rgb]{ .851,  .882,  .957}45.74  & \cellcolor[rgb]{ .851,  .882,  .957}54.01  & \cellcolor[rgb]{ .851,  .882,  .957}35.20  & \cellcolor[rgb]{ .851,  .882,  .957}43.51  \\
\cmidrule{2-11}          & \multicolumn{1}{c}{\multirow{2}[1]{*}{W+F}} & \multicolumn{1}{c}{Mix-PT} & 28.75  & 35.00  & 41.25  & 54.17  & 49.79  & 52.97  & 32.40  & 42.05  \\
          & \multicolumn{1}{c}{} & \multicolumn{1}{c}{\cellcolor[rgb]{ .851,  .882,  .957}CrossIC-PT} & \cellcolor[rgb]{ .851,  .882,  .957}31.25  & \cellcolor[rgb]{ .851,  .882,  .957}38.33  & \cellcolor[rgb]{ .851,  .882,  .957}40.75  & \cellcolor[rgb]{ .851,  .882,  .957}58.33  & \cellcolor[rgb]{ .851,  .882,  .957}49.87  & \cellcolor[rgb]{ .851,  .882,  .957}53.90  & \cellcolor[rgb]{ .851,  .882,  .957}33.60  & \cellcolor[rgb]{ .851,  .882,  .957}43.72  \\
    \midrule
    \midrule
    \multirow{3}[0]{*}{Qwen-2.5-7B} & \multicolumn{2}{c}{base} & 35.00  & 61.67  & 46.50  & 60.83  & 60.14  & 53.73  & 58.80  & 53.81  \\
          & \multicolumn{2}{c}{Cross-CPT} & 36.25  & 60.83  & 47.25  & 70.83  & 64.37  & 53.77  & 60.00  & 56.19  \\
          & \multicolumn{2}{c}{\cellcolor[rgb]{ .851,  .882,  .957}CrossIC-CPT} & \cellcolor[rgb]{ .851,  .882,  .957}35.00  & \cellcolor[rgb]{ .851,  .882,  .957}62.50  & \cellcolor[rgb]{ .851,  .882,  .957}48.25  & \cellcolor[rgb]{ .851,  .882,  .957}70.00  & \cellcolor[rgb]{ .851,  .882,  .957}69.87  & \cellcolor[rgb]{ .851,  .882,  .957}55.66  & \cellcolor[rgb]{ .851,  .882,  .957}60.00  & \cellcolor[rgb]{ .851,  .882,  .957}57.33  \\
    \midrule
    \midrule
    \multirow{3}[1]{*}{Qwen-2.5-1.5B} & \multicolumn{2}{c}{base} & 32.50  & 35.83  & 35.00  & 49.17  & 65.49  & 42.22  & 29.60  & 41.40  \\
          & \multicolumn{2}{c}{Cross-CPT} & 36.25  & 35.83  & 38.50  & 51.67  & 61.52  & 37.84  & 24.80  & 40.92  \\
          & \multicolumn{2}{c}{\cellcolor[rgb]{ .851,  .882,  .957}CrossIC-CPT} & \cellcolor[rgb]{ .851,  .882,  .957}37.50  & \cellcolor[rgb]{ .851,  .882,  .957}33.33  & \cellcolor[rgb]{ .851,  .882,  .957}38.75  & \cellcolor[rgb]{ .851,  .882,  .957}54.17  & \cellcolor[rgb]{ .851,  .882,  .957}64.05  & \cellcolor[rgb]{ .851,  .882,  .957}41.83  & \cellcolor[rgb]{ .851,  .882,  .957}26.00  & \cellcolor[rgb]{ .851,  .882,  .957}42.23  \\
    \bottomrule
    \end{tabular}}%
    
    \caption{The results of our method, ablation study, and the baselines in Korean, Porturguese and Thai.}
  \label{task_results1}%
\end{table}%

\section{Results in Other Non-Latin Languages}
To further validate the generalization capability of CrossIC-PT, we conducted additional experiments on Chinese and Vietnamese. As shown in Table~\ref{zh_vi}, our method demonstrates consistent effectiveness across these non-Latin script languages.

\begin{table}[htbp]
  \centering
    \resizebox{0.85\textwidth}{!}{\begin{tabular}{c|c|c|c|c|c|c|c|c}
    \toprule
        \multicolumn{1}{c|}{\multirow{2}[2]{*}{\textbf{Chinese}}} & \multicolumn{7}{c}{\textbf{Tasks}}     & \multicolumn{1}{|c}{\multirow{2}[2]{*}{\textbf{AVG}}} \\
      & \multicolumn{1}{c}{\textbf{XLOGIQA}} & \multicolumn{1}{c}{\textbf{XHELLASWAG}} & \multicolumn{1}{c}{\textbf{MMMLU}} & \multicolumn{1}{c}{\textbf{XNLI}} & \multicolumn{1}{c}{\textbf{MRC}} & \multicolumn{1}{c}{\textbf{FLORES-200}} & \multicolumn{1}{c|}{\textbf{MGSM}} &  \\
    \midrule
    Llama-3.1-8B & \textbf{48.75 } & 35.09  & 35.75  & 56.67  & 54.60  & 35.67  & 38.00  & 43.50  \\
    \midrule
    Mix-PT & 43.75  & 32.46  & 40.25  & \textbf{60.83 } & 60.34  & 36.17  & 36.00  & 44.26  \\
    \rowcolor[rgb]{ .851,  .882,  .957} CrossIC-PT & 45.00  & \textbf{36.97 } & \textbf{41.00 } & \textbf{60.83 } & \textbf{61.77 } & \textbf{37.37 } & \textbf{39.60 } & \textbf{46.08 } \\
    \midrule
    \multicolumn{1}{r}{} & \multicolumn{1}{r}{} & \multicolumn{1}{r}{} & \multicolumn{1}{r}{} & \multicolumn{1}{r}{} & \multicolumn{1}{r}{} & \multicolumn{1}{r}{} & \multicolumn{1}{r}{} & \multicolumn{1}{r}{} \\

    \midrule    
     \multicolumn{1}{c|}{\multirow{2}[2]{*}{\textbf{Vietnamese}}} & \multicolumn{7}{c}{\textbf{Tasks}}     & \multicolumn{1}{|c}{\multirow{2}[2]{*}{\textbf{AVG}}} \\
     & \multicolumn{1}{c}{\textbf{XLOGIQA}} & \multicolumn{1}{c}{\textbf{XHELLASWAG}} & \multicolumn{1}{c}{\textbf{MMMLU}} & \multicolumn{1}{c}{\textbf{XNLI}} & \multicolumn{1}{c}{\textbf{MRC}} & \multicolumn{1}{c}{\textbf{FLORES-200}} & \multicolumn{1}{c|}{\textbf{MGSM}} &  \\
    \midrule
    Llama-3.1-8B & \textbf{46.25 } & 37.50  & 41.75  & 55.83  & 51.31  & 36.69  & 36.80  & 43.73  \\
    \midrule
    Mix-PT & 43.75  & 39.17  & 42.75  & 60.00  & 47.43  & 36.75  & \textbf{39.60 } & 44.21  \\
    \rowcolor[rgb]{ .851,  .882,  .957} CrossIC-PT & 43.75  & \textbf{41.38 } & \textbf{43.50 } & \textbf{62.50 } & \textbf{58.73 } & \textbf{40.09 } & 39.20  & \textbf{47.02 } \\
    \bottomrule
    \end{tabular}}%
  \caption{The results of CrossIC-PT in Chinese and Vietnamese.}
  \label{zh_vi}%
\end{table}%

\section{Results in Instruction-Tuned LLMs}
To comprehensively evaluate our approach, we extended experiments to instruction-tuned models using Llama-3-8B-Instruct for Thai. We maintained all default parameters except for setting the learning rate to 5e-6. As shown in Table~\ref{instruct}, CrossIC-PT demonstrates performance limitations on Thai XLOGIQA, XHELLASWAG, and XNLI tasks, while maintaining advantages in knowledge-based (MMMLU), comprehension (MRC), and translation (FLORES-200) benchmarks.

These results suggest that while CrossIC-PT remains effective for certain capabilities, its performance on instruction-tuned models reveals limitations potentially attributable to the divergence between fine-tuning and pre-training objectives. This finding indicates the need for further investigation into optimal methods for leveraging semantically related contextual corpora in instruction-tuned settings.

\begin{table}[htbp]
  \centering
    \resizebox{0.85\textwidth}{!}{\begin{tabular}{c|ccccccc|c}
    \toprule
    \multicolumn{1}{c|}{\multirow{2}[2]{*}{\textbf{Thai}}} & \multicolumn{7}{c}{\textbf{Tasks}}     & \multicolumn{1}{|c}{\multirow{2}[2]{*}{\textbf{AVG}}} \\
      & \multicolumn{1}{c}{\textbf{XLOGIQA}} & \multicolumn{1}{c}{\textbf{XHELLASWAG}} & \multicolumn{1}{c}{\textbf{MMMLU}} & \multicolumn{1}{c}{\textbf{XNLI}} & \multicolumn{1}{c}{\textbf{MRC}} & \multicolumn{1}{c}{\textbf{FLORES-200}} & \multicolumn{1}{c|}{\textbf{MGSM}} &  \\
    \midrule
    Llama-3.1-8B-Instruct & \textbf{37.50 } & 35.00  & 42.25  & \textbf{59.17 } & 64.47  & 50.86  & \textbf{47.60 } & 48.12  \\
    Mix-PT & 30.00  & \textbf{37.50 } & 44.75  & 54.17  & 68.78  & 53.46  & 46.40  & 47.87  \\
    \rowcolor[rgb]{ .851,  .882,  .957} CrossIC-PT & 36.25  & 35.00  & \textbf{45.00 } & 58.33  & \textbf{69.96 } & \textbf{54.25 } & 47.20  & 49.43  \\
    \bottomrule
    \end{tabular}}%
  \caption{The results of CrossIC-PT in Thai based on Llama-3.1-8B-Instruct.}
  \label{instruct}%
\end{table}%

\section{Extended Context Window Analysis}

To thoroughly investigate the impact of context window size on model performance, we experimented with an extended context window. Specifically, we evaluated the Qwen-2.5-1.5B model on the Thai language by increasing the context window length to 8192 tokens. The results, presented in Table \ref{tab:extended_window_thai_appendix}, indicate that employing an 8192-token window yielded further performance improvements. This configuration achieved an additional 1.33\% gain compared to the 4096-token CrossIC-PT model, underscoring the benefits of a larger context window for processing extensive bilingual input.

\begin{table}[h!]
\centering
\label{tab:extended_window_thai_appendix}
\resizebox{0.85\textwidth}{!}{\begin{tabular}{l c c c c c c c c}
\toprule
\textbf{Model} & \textbf{XLOGIQA} & \textbf{XHELLASWAG} & \textbf{MMMLU} & \textbf{XNLI} & \textbf{MRC} & \textbf{FLORES-200} & \textbf{MGSM} & \textbf{AVG.} \\
\midrule
Qwen2.5-1.5B & 32.50 & \textbf{35.83} & 35.00 & 49.17 & 65.49 & \textbf{42.22} & 29.60 & 41.40 \\
Mix-PT & 36.25 & \textbf{35.83} & 38.50 & 51.67 & 61.52 & 37.84 & 24.80 & 40.92 \\
CrossIC-PT window\_length=4096 & 37.50 & 33.33 & 38.75 & \textbf{54.17} & 64.05 & 41.83 & 26.00 & 42.23 \\
CrossIC-PT window\_length=8192 & \textbf{38.75} & 32.50 & \textbf{41.00} & 51.67 & \textbf{66.84} & 41.77 & \textbf{32.40} & \textbf{43.56} \\
\bottomrule
\end{tabular}}
\caption{Performance on Thai with Extended Context Window}
\end{table}

This expanded experimental setup provides valuable insights into the architectural considerations for handling long-range dependencies in multilingual contexts.

\end{document}